\documentclass[11pt]{article}

\usepackage[final]{acl}

\usepackage{times}
\usepackage{latexsym}

\usepackage[T1]{fontenc}

\usepackage[utf8]{inputenc}

\usepackage{microtype}

\usepackage{inconsolata}

\usepackage{graphicx}

\newcommand{\eat}[1]{}

\usepackage{amsfonts}
\usepackage{colortbl}
\usepackage{xspace}
\usepackage{comment}
\usepackage{booktabs}
\usepackage{multirow}
\usepackage{framed}
\usepackage{subcaption}

\usepackage{circledsteps}

\definecolor{green}{RGB}{0,128,0}

\definecolor{yellow}{RGB}{255,200,18}

\newcommand{\add}[1]{\textcolor{black}{{#1}}}

\newcommand{\stab}{\vspace{1.2ex}\noindent}

\newcommand{\bi}{\begin{itemize}}
\newcommand{\ei}{\end{itemize}}

\newcommand{\be}{\begin{enumerate}}
\newcommand{\ee}{\end{enumerate}}
\newcommand{\beqn}{\begin{eqnarray*}}
\newcommand{\eeqn}{\end{eqnarray*}}

\newcommand{\stitle}[1]{\stab\noindent{\bf #1}}

\newcommand{\eg}{{\em e.g.,}\xspace}

\usepackage{listings}

\usepackage{pifont}
\usepackage{tikz}
\usepackage{enumitem}

\usepackage{tcolorbox}
\tcbuselibrary{skins, breakable, theorems}

\usepackage{placeins}
\usepackage{float}

\newcommand{\sys}{{\textbf{TradeLens}}\xspace}

\definecolor{yy}{HTML}{f0c38e}
\definecolor{rr}{HTML}{f38181}

\begin{document}

\title{Can Agentic Trading Systems Pay for Their Own Intelligence?}
% Just set the title in a qustion tone, and drive reader to find more in abstract

% Author information using the modern ACL multi-author style.
% In review mode (\usepackage[review]{acl}), this block is anonymized automatically.
\author{
  \textbf{Qiqi Duan}\textsuperscript{1$\ast$},
  \textbf{Changlun Li}\textsuperscript{1,2$\ast$},
  \textbf{Chen Wang}\textsuperscript{1$\ast$},
  \textbf{Fan Zhang}\textsuperscript{2,4},\\
  \textbf{Mengxiang Wang}\textsuperscript{1},
  \textbf{Dayi Miao}\textsuperscript{1},
  \textbf{Peixian Ma}\textsuperscript{2},
  \textbf{Jiangpeng Yan}\textsuperscript{3},\\
  \textbf{Liyuan Chen}\textsuperscript{3},
  \textbf{Shuoling Liu}\textsuperscript{3},
  \textbf{Preslav Nakov}\textsuperscript{4},
  \textbf{Yuyu Luo}\textsuperscript{1,2},
  \textbf{Nan Tang}\textsuperscript{1,2$\dagger$}
\\[4pt]
  \textsuperscript{1}HKUST(GZ) \quad
  \textsuperscript{2}Paradoox AI Research\quad
  \textsuperscript{3}E Fund Management Co., Ltd \quad
  \textsuperscript{4}MBZUAI \quad
\\[2pt]
  \small{$\ast$ Equal contribution. \quad
         $\dagger$ Corresponding author: \texttt{nantang@hkust-gz.edu.cn}}
}

\maketitle

\begin{abstract}

Large language model (LLM) agents are increasingly used in trading systems, where model reasoning, tool use, and continual decisions incur costs that are expected to produce trading value. Existing evaluations typically report performance metrics, but rarely examine \textbf{\emph{agentic viability}}: whether dynamic LLM-mediated decisions convert their induced costs into measurable incremental profit.
To apply this criterion, we introduce \sys, a trace-grounded diagnostic toolkit for evaluating agentic trading systems from their trading records, runtime traces, and deployment configurations. It reconstructs trading trajectories, attributes profit and cost to interpretable evidence, and diagnoses whether and why an agent pays for its own intelligence.
We conduct extensive analysis across backbone models, capital scales, trading frequencies, and system architectures, together with deployment discussion. 
Our results show that viability hinges on intelligence-to-profit conversion: models exhibit different failure patterns, such as poor asset selection in DeepSeek-V3.2 and negative timing in GLM-4.7, while capital scale, trading frequency, and architecture 
% matter only by amplifying or degrading decision-attributed timing value.
\add{ matter primarily in these runs, especially through their effects on decision-attributed timing value.}
These findings reframe the evaluation of LLM-based trading agents from capability-centric performance ranking to trace-grounded diagnosis of intelligence-to-profit conversion.
Our code is available at \url{https://github.com/ParadooxAI/TradeLens}.

\end{abstract}

\section{Introduction}

\begin{figure}[t!]
    \centering
    \includegraphics[width=\linewidth]{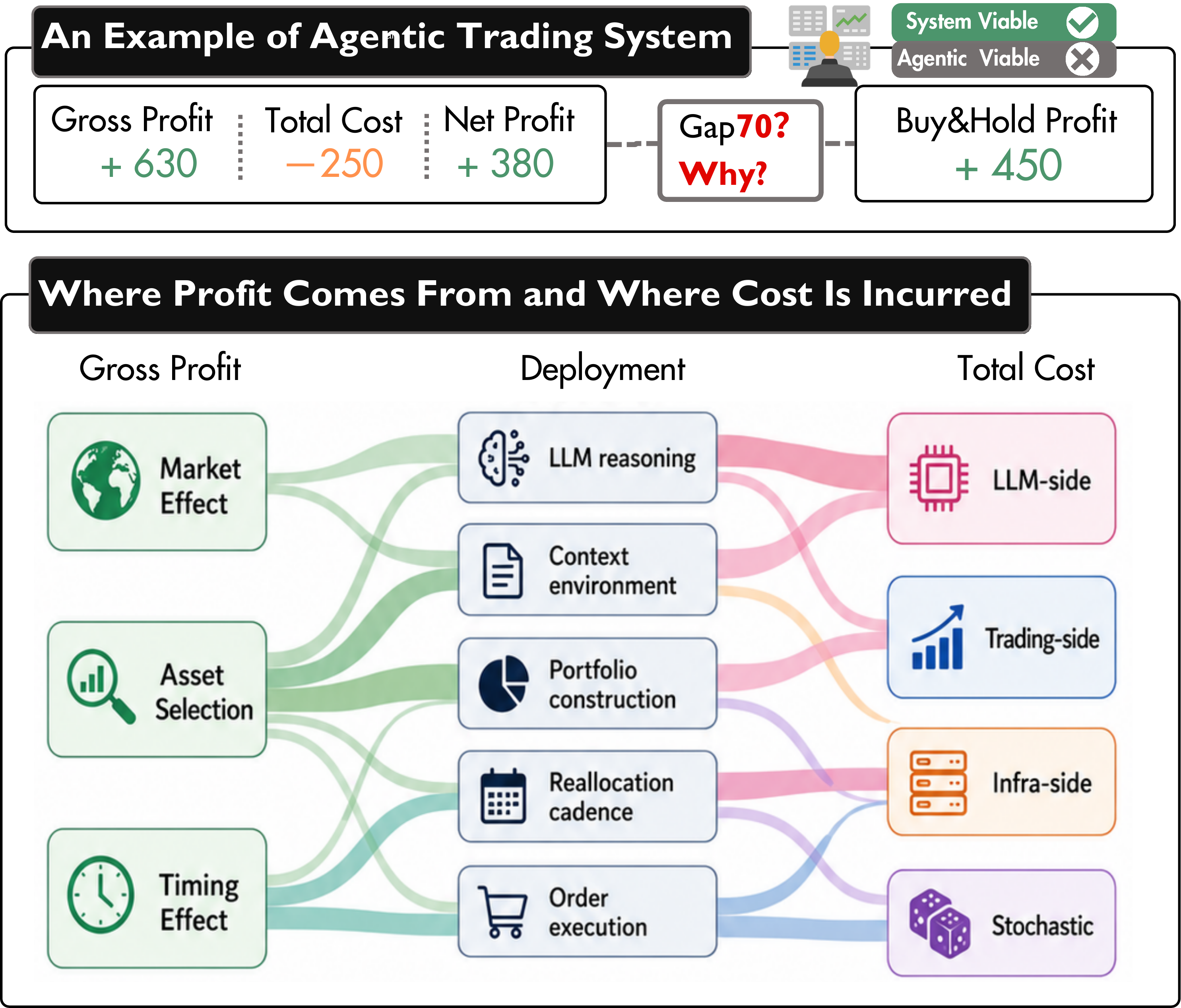}
    \caption{
    % Motivation. A profitable agentic trading system may still fail to create useful trading value, and diagnose ``Why'' because both returns and costs arise from intertwined deployment drivers.
    \add{Motivation. Profitability does not necessarily imply positive agentic value. Our framework diagnoses why by disentangling the deployment drivers that jointly shape returns and costs.}}
    \label{fig:motivation}
\end{figure}

With strong general reasoning abilities~\cite{yu2023metamath,liu2023is,guo2025deepseek} and increasing adaptation to financial tasks~\cite{wu2023bloomberggpt,chen2025advancing,liu2023fingpt,zhang2026beyond,liu2025deepseek}, LLM agents are moving from financial question answering to direct participation in trading workflows.

Recent systems use LLMs as trading copilots~\cite{zhou2024finrobot,guo2025large}, portfolio managers~\cite{zhao2025alphaagents,nextfund}, and market-analysis agents equipped with external tools~\cite{han2025finsphere,zhang2026alpha,zhang2026finreporting}. 

\stitle{Observation.} 
Despite promising results, a critical gap remains in how agentic trading systems are evaluated. 
Most existing studies report \emph{gross profit} or metrics such as Sharpe ratio~\cite{li2025investorbench,chen2025stockbench,xie2023pixiu}. 
While useful for measuring trading performance, these metrics are incomplete for deployment: developers must assess not only how much profit a system makes, but also where the profit comes from and how deployment costs affect the realized margin. 
This echoes recent cost-aware LLM evaluations balancing monetary costs and reasoning efficiency~\cite{erol2025cost,zellinger2025economic,wang2025efficient,zhang2024cut},
\add{and whether AI expenditure ultimately yields measurable economic value~\cite{howell2023economic,nanda2025state}.}

However, \add{this question becomes more stringent in agentic way: cost} is the price paid for LLM-mediated intelligence, including model reasoning, tool use, memory retrieval, and continual decision-making. 
Beyond operating within a budget~\cite{wang2024reasoning}, the system should also meet the \emph{token-economy requirement}~\cite{chen2026token}: generating enough incremental profit to justify the intelligence consumed by its trading decisions.

\stitle{The blind spot.}
This requirement creates a profit-side and cost-side double blind in evaluating agentic trading systems. 
As shown in Figure~\ref{fig:motivation}, an agentic system may earn a gross profit of 630, but after 250 in deployment costs, its net profit drops to 380; meanwhile, a simple buy-and-hold baseline without LLM inference or active trading costs may earn 450 over the same period. 
Thus, gross profit can overstate decision value when the agent does not outperform passive market exposure, and can overstate deployability when end-to-end costs absorb the realized margin. 
We therefore ask: \emph{Can agentic trading systems pay for their own intelligence, and how can we diagnose whether LLM-mediated decisions convert their induced costs into incremental trading value?}

\stitle{Call for rigorous assessment.}
% Answering this question requires explicit profit-cost analysis. 
% As illustrated in Figure~\ref{fig:motivation}, an agentic trading system cannot be evaluated from a single final return because both profit and cost are shaped by the agent's deployment process. 
% A profitable trajectory may mainly reflect passive market exposure or favorable initial asset selection, while the agent's dynamic decisions contribute little or even reduce value. 
% Conversely, higher reasoning, tool-use, or execution costs are not necessarily wasteful if they produce sufficient active profit. 
% Thus, before diagnosing whether an agent pays for its own intelligence, we must first measure where profit comes from and where cost is incurred.
%
Answering this question requires identifying both the value added by LLM intelligence and the costs it incurs.
The decomposition in the lower panel of Figure~\ref{fig:motivation} shows that both profit and cost are shaped by the deployment process.
A profitable trajectory may mainly reflect market movement or favorable initial asset selection, while the agent's dynamic decisions contribute little or even reduce value. 
Conversely, higher reasoning, tool-use, or execution costs are not necessarily wasteful if they produce sufficient active profit. 
This makes rigorous assessment necessary for diagnosing whether LLM-mediated decisions are economically justified.
% To assess whether an agentic system is worth deploying, we first need to identify where profit comes from and where cost is incurred.
% A rigorous assessment should therefore ask not only how much profit is realized, but also which part is attributable to LLM-mediated decisions and whether that value exceeds the costs they introduce.

\stitle{Challenge of trace-grounded diagnosis.}
However, performing this diagnosis is difficult because the evidence needed to evaluate LLM-mediated decisions is fragmented across different records.
Portfolio trajectories show what the system earned, but not whether the gain came from passive exposure, initial allocation, or dynamic agent intervention. 
Runtime traces show what the agent did, including model calls, context processing, tool use, retries, and trading actions, but not whether these activities improved asset selection, timing, or portfolio value. 
Deployment costs further vary with model choice, context length, tool-use frequency, trading cadence, and execution behavior, making a single cost proxy insufficient. 
Therefore, the challenge is to ground the diagnosis of intelligence-to-profit conversion in heterogeneous evidence from trading outcomes, runtime traces, and deployment costs.

\stitle{Solution.}
To address these challenges, we propose \sys, a trace-grounded diagnosis toolkit for evaluating whether agentic trading systems can pay for their own intelligence. 
This toolkit first reconstructs the investment process from portfolio records, runtime traces, and deployment configurations, and attributes profit and cost to interpretable components. 
Based on this evidence, we define two viability notions~(Section~\ref{sec:method}): \emph{system viability}, which asks whether the deployed system pays for itself after end-to-end costs, and \emph{agentic viability}, which asks whether dynamic LLM-mediated decisions create enough active value to justify their decision-induced costs. 
% \sys then invokes an LLM-based diagnosis agent to generate evidence-grounded explanations and strategy-level suggestions from the observed trading trajectory, and converts these diagnoses into visual and textual reports~(Section~\ref{sec:toolkit}).
% % 
Then, an LLM-based diagnosis agent generates evidence-grounded explanations and strategy-level suggestions from the observed trading trajectory, and converts these diagnoses into visual and textual reports~(Section~\ref{sec:toolkit}).
% We evaluate \sys through controlled experiments and practitioner feedback, showing that it reveals failure modes hidden by gross profit alone and helps users assess deployability and revision directions~(Section~\ref{sec:exp}).
We evaluate \sys through controlled experiments and practitioner feedback, showing that it reveals failure modes hidden by system's profitability and helps users assess deployability and identify revision directions~(Section~\ref{sec:exp}).

% \stitle{Contributions.}
Collectively, we reframe the evaluation of agentic trading systems from capability-centric to economically grounded. To summarize, we make three contributions: 

% (i) we formulate profit--cost viability for agentic trading systems, decomposing gross profit and deployment cost while separating whole-system profitability from the economic value of dynamic agentic intervention;

(i) We formulate \emph{profit--cost viability} as a trace-grounded evaluation problem for agentic trading systems, distinguishing whether the overall deployed system pays for itself and whether LLM-mediated decisions justify their induced costs.

(ii) We build \sys, an open-source audit toolkit that operationalizes the methodology using trading records, runtime traces, and deployment configurations, and produces evidence-grounded diagnostic reports.

(iii) We conduct empirical viability studies across backbone model, capital scale, trading frequency, and architecture, together with practitioner feedback, showing how profit sources and intelligence cost jointly determine deployability.

% Collectively, we reframe the evaluation of agentic trading systems from capability-centric to economically grounded.

\section{Related Work}
\label{sec:related-work}

\subsection{LLM-Based Trading Agents}

LLM-based trading agents have expanded from financial analysis and signal generation~\cite{wang2025financial,xing2025designing} to decision pipelines that use memory, tools, and multi-agent coordination~\cite{yu2024fincon,yu2025finmem,xiao2025tradingagents,Li2025HedgeAgentsAB}. Recent work also moves evaluation closer to practical trading settings, including cryptocurrency markets~\cite{li2024cryptotrade,luo2025llmpowered}, portfolio construction~\cite{guo2025mass,zhao2025alphaagents}, streaming interaction~\cite{li2025time,fan2025aitrader}, and end-to-end research platforms~\cite{suntrademaster,zhang2026finworld}.
%
% Across these settings, evaluation commonly emphasizes capability, prediction quality, or gross trading performance. This leaves a separate question underexplored: whether the observed trading gains remain meaningful after accounting for profit sources and deployment costs. \sys addresses this evaluation question as a diagnosis toolkit for existing trading agents, rather than as another trading architecture.
Across these works, evaluation mainly focuses on capability or gross trading performance, leaving profit sources and economic viability underexplored. 
\sys addresses this gap as a diagnosis toolkit for existing trading agents, checking whether these agents can pay for their intelligence.

\subsection{Cost-Aware and Trace-Grounded AI Evaluation}

Recent work increasingly incorporates computational cost into model evaluation. 
Cost-of-pass estimates the monetary cost of obtaining a successful answer~\cite{erol2025cost,zellinger2025economic}; efficient-reasoning studies examine thinking style and redundancy in agentic systems~\cite{wang2025harnessing,lin2025plan,wang2025efficient,zhang2024cut}; and other work advances resource-constrained evaluation~\cite{chen2024frugalgpt,huang2025thriftllm,liu2025costbench,yang2025bamas,datamosaic,datamosaic_demo}.
These studies mainly treat cost as a resource-efficiency variable. 
For LLM agents, however, costs are also induced by the decision process itself, including reasoning, tool use, memory access, and multi-step interaction~\cite{shinn2023reflexion}.
Token economics further frames these activities as paid token and compute consumption before downstream actions are produced~\cite{wang2024reasoning,chen2026token}.
Prior work on agent evaluation and diagnosis has shown that such trajectories provide useful evidence beyond final task outcomes~\cite{he2026traject,ou2025agentdiagnose}. 
In trading workflows, we use this trace-grounded view in a different setting: agent activities can be aligned with resource use, trading actions, and portfolio outcomes, enabling diagnosis of whether decision-induced costs are converted into trading value.

\subsection{Performance Attribution in Financial Evaluation}

Financial evaluation has long separated raw return from interpretable sources of return. Risk-adjusted metrics such as the Sharpe ratio~\cite{sharpe1994sharpe} and regime-aware analyses~\cite{ang2012regime} show that portfolio outcomes can reflect volatility or market conditions rather than decision skill. Classical performance attribution decomposes benchmark-relative returns into allocation, selection, and interaction effects~\cite{brinson1986determinants,brinson1991determinants}, with later work using attribution to study active management skill and dynamic allocation behavior~\cite{grinold2000active,hsu2010performance,al2018outperformance}.

Recent financial AI benchmarks and trading-agent studies often report predictive accuracy, cumulative return, or portfolio-level profit~\cite{ko2024can,li2025investorbench,chen2025stockbench,fan2025aitrader}. These metrics are useful, but they do not explain whether profit comes from market exposure, asset selection, or dynamic reallocation. \sys adopts a lightweight attribution view for this purpose and joins it with system-cost accounting, so viability is assessed from decision-attributed margin rather than gross profit alone.

% Trace-Grounded Profit–Cost Diagnosis
\section{Methodology: Profit--Cost Viability}
\label{sec:method}

This section defines how we evaluate whether an agentic trading system can pay for the intelligence it consumes.
We focus on two questions: whether the system is economically viable, and whether its agentic decisions generate sufficient incremental value to justify their induced costs.

\subsection{Profit and Cost Attribution}
\label{sec:measurement}

Given an evaluation window \([1,T]\), let \(V_0\) be the initial portfolio value and \(V_T^{\mathrm{dyn}}\) be the realized terminal value of the deployed agentic system.
The cumulative gross profit is

\begin{equation}
    P_{1:T}=V_T^{\mathrm{dyn}}-V_0 .
    \label{eq:gross-profit}
\end{equation}

\stitle{Profit Attribution.}
Classical performance attribution decomposes portfolio returns into interpretable sources before evaluating investment decisions~\cite{brinson1986determinants,brinson1991determinants}. 
We adopt this logic for agentic trading systems, but reformulate it at the portfolio-value level.
%
% For diagnosis, we further separate gross profit into three components.
% Let \(V_T^{\mathrm{sys}}\) denote the terminal value of a passive market benchmark, and let \(V_T^{\mathrm{base}}\) denote the terminal value obtained by holding the system's initial allocation unchanged.
% We define
%
% For diagnosis, we separate gross profit using two nested counterfactual baselines over the same asset universe and evaluation window. 
% %
% Let \(V_T^{\mathrm{sys}}\) denote the terminal value of a passive market benchmark, and let \(V_T^{\mathrm{base}}\) denote the terminal value obtained by holding the system's initial allocation unchanged. 
% The comparison between these baselines separates aggregate market exposure from initial asset-selection value, while the remaining difference between the realized dynamic path and the static-allocation path is attributed to timing.
For diagnosis, we separate gross profit using two counterfactual baselines over the same asset universe and evaluation.

Let \(V_T^{\mathrm{sys}}\) denote the terminal value of a passive market benchmark based on the system's initial market exposure, and let \(V_T^{\mathrm{base}}\) denote the terminal value obtained by holding the system's first active allocation unchanged.
The comparison between these baselines separates market exposure from asset-selection value, while the remaining difference between the realized dynamic path and the static-allocation path is attributed to timing:

\begin{equation}
\begin{aligned}
    P^{\mathrm{sys}} &= V_T^{\mathrm{sys}} - V_0,\\
    P^{\mathrm{asset}} &= V_T^{\mathrm{base}} - V_T^{\mathrm{sys}},\\
    P^{\mathrm{timing}} &= V_T^{\mathrm{dyn}} - V_T^{\mathrm{base}}.
\end{aligned}
    \label{eq:profit-components}
\end{equation}

Thus,
\begin{equation}
    P_{1:T}
    =
    P^{\mathrm{sys}}
    +
    P^{\mathrm{asset}}
    +
    P^{\mathrm{timing}} .
    \label{eq:profit-decomp}
\end{equation}

% Here, \(P^{\mathrm{sys}}\) captures broad market exposure, \(P^{\mathrm{asset}}\) captures the effect of the initial allocation relative to the market benchmark, and \(P^{\mathrm{timing}}\) captures the incremental value of subsequent dynamic reallocation.
Here, \(P^{\mathrm{sys}}\), \(P^{\mathrm{asset}}\), and \(P^{\mathrm{timing}}\) correspond to market exposure, asset selection, and timing decisions, respectively. 
In our analysis, \(P^{\mathrm{timing}}\) is treated as the profit component most directly attributable to agentic intervention.

\stitle{Cost Attribution.}
Transaction cost economics treats market participation as non-frictionless: information processing, decision-making, coordination, and execution all consume resources~\cite{Williamson2005}. For agentic trading systems, ``intelligence'' is produced through paid inference and then realized through execution and deployment.
We decompose the deployment cost in each period \(t\) as
\begin{equation}
    C_t
    =
    C_t^{\mathrm{llm}}
    +
    C_t^{\mathrm{trd}}
    +
    C_t^{\mathrm{inf}}
    +
    C_t^{\mathrm{sto}},
    \label{eq:cost-decomp}
\end{equation}

\noindent where \(C_t^{\mathrm{llm}}\) is LLM inference cost, \(C_t^{\mathrm{trd}}\) is trading execution cost, \(C_t^{\mathrm{inf}}\) is infrastructure and data-access cost, and \(C_t^{\mathrm{sto}}\) captures residual hard-to-enumerate costs.
The cumulative cost is
\begin{equation}
    C_{1:T}=\sum_{t=1}^{T} C_t .
    \label{eq:total-cost}
\end{equation}

Detailed profit and cost itemization is provided in Appendix~\ref{app:profit} and Appendix~\ref{app:cost}.

\subsection{Viability Criteria}
\label{sec:viability}

Viability is evaluated over a common decision window \([1,T]\). 
We then distinguish two levels of viability. 
System viability asks whether the whole pipeline pays for itself, whereas agentic viability asks a more specific agent-evaluation question: whether dynamic LLM intervention creates enough incremental value to justify the costs it induces.

\stitle{System viability.}
System viability asks whether the fully deployed pipeline pays for itself.
The system-level net profit is

\begin{equation}
    R_{1:T}=P_{1:T}-C_{1:T}.
    \label{eq:net-profit}
\end{equation}

The system is viable if

\begin{equation}
    \mathrm{SystemViable}
    =
    \mathbb{I}\!\left[
    R_{1:T} \ge 0
    \right].
    \label{eq:system-viable}
\end{equation}

\stitle{Agentic viability.}
Agentic viability asks whether dynamic agent intervention itself is economically justified.
It removes passive market exposure (\(P^{\mathrm{sys}}\)) and initial asset-selection effects (\(P^{\mathrm{asset}}\)), and compares timing profit against costs that are directly induced by dynamic decisions. 
% We compare the timing profit with the costs directly induced by agent activity.
%
We classify costs that scale with token usage, retries, and tool calls as dynamic costs, and treat fixed hosting as static costs.
The dynamic cost is

\begin{equation}
    C_{1:T}^{\mathrm{dyn}}
    =
    \sum_{t=1}^{T}
    \left(
    C_t^{\mathrm{llm}}
    +
    C_t^{\mathrm{trd}}
    +
    C_t^{\mathrm{sto}}
    \right),
    \label{eq:dyn-cost}
\end{equation}

% excluding infrastructure cost that does not vary with agent decisions.
excluding infrastructure cost which are treated as system-level deployment costs.
The net agentic value is

\begin{equation}
    R_{1:T}^{\mathrm{agent}}
    =
    P^{\mathrm{timing}}
    -
    C_{1:T}^{\mathrm{dyn}} .
    \label{eq:agentic-net}
\end{equation}

The agent is viable if

\begin{equation}
    \mathrm{AgenticViable}
    =
    \mathbb{I}\!\left[
    R_{1:T}^{\mathrm{agent}} \ge 0
    \right].
    \label{eq:agentic-viable}
\end{equation}

\section{TradeLens}
\label{sec:toolkit}

\begin{figure*}
    \centering
     \includegraphics[width=0.85\linewidth]{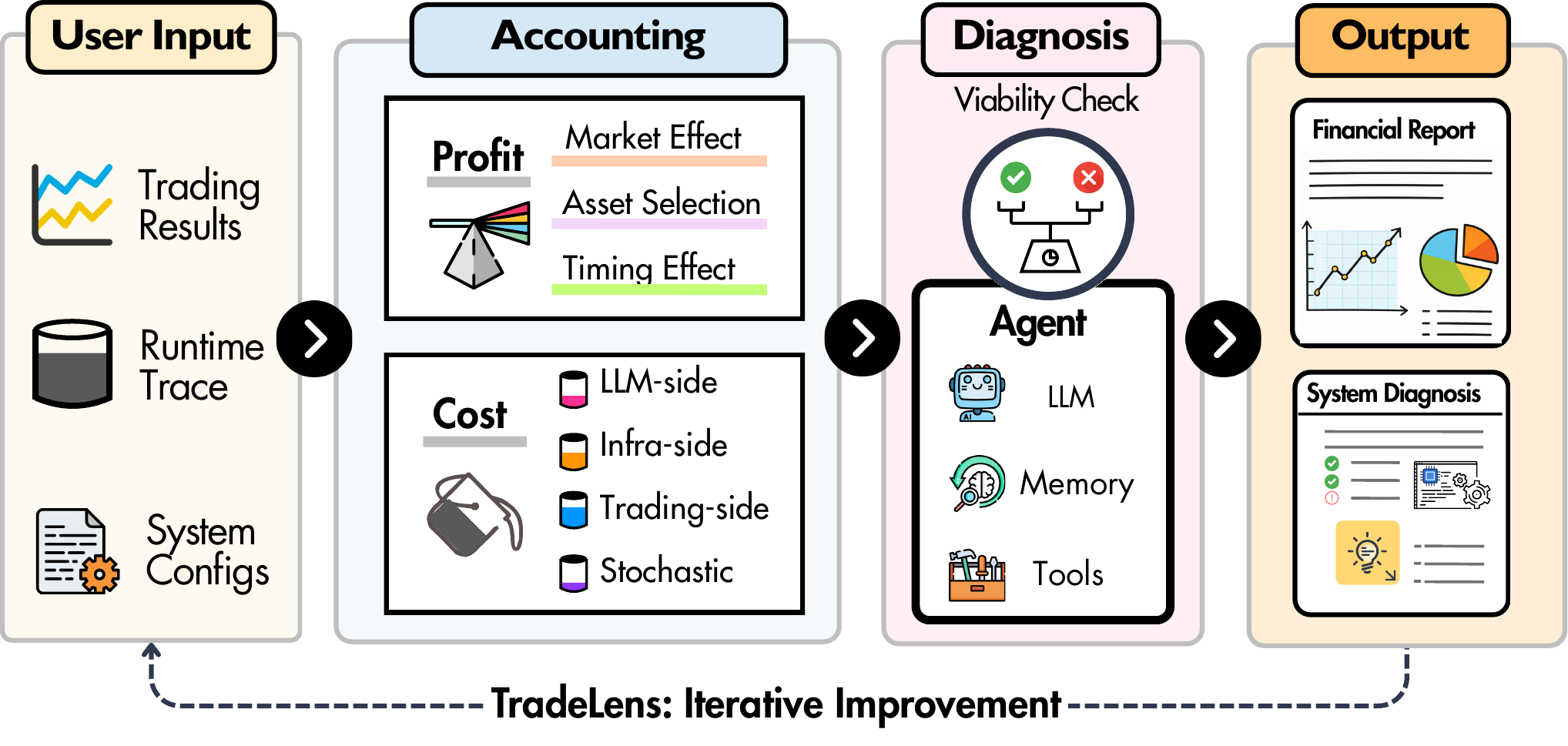}
    \caption{\textbf{The pipeline overview.}
    % Retail traders can utilize the toolkit to evaluate the profit--cost viability of their own agentic systems under realistic assumptions by providing trading results, runtime traces, and basic configurations. Consequently, they can gain iterative improvements and achieve promising results.
    \add{Retail traders can use the toolkit to evaluate the profit–cost viability of their agentic systems under realistic assumptions using trading results, runtime traces, and basic configurations. The diagnostics support an iterative loop of system adjustment and re-evaluation.}
    }
    \label{fig:framework}
\end{figure*}

In this section, we describe how \sys implements this methodology by reconstructing trading trajectories, aggregating runtime costs, and generating structured diagnostic reports.
% As shown in Figure~\ref{fig:framework}, \sys is designed as an intermediate layer on an agentic trading system.

\subsection{Overview}
\label{sec:toolkit-overview}

The design goal of \sys is to make agentic trading systems auditable without assuming a specific agent architecture.
Different systems may use different LLM backbones, prompting strategies, tool chains, memory modules, or multi-agent designs.
However, once deployed, they tend to leave similar observable traces.
\sys therefore defines a trace interface around the following three inputs: \emph{trading results}, \emph{runtime traces}, and \emph{system configurations}.

\stitle{Input.}
Trading results describe what happened in the market-facing process, including portfolio values, position weights, executed orders, and transaction logs, used to reconstruct the realized portfolio trajectory.
Runtime traces describe how the system produced its decisions, including model usage, tool calls, retries, latency, and execution events, used to recover cost drivers and system activities.
System configurations describe the deployment setting, including the LLM backbone, capital scale, trading frequency, market window, cost assumptions, and agent architecture.
They provide the context needed to interpret both profit and cost.

% This interface makes \sys architecture-agnostic.
% The toolkit does not require access to the internal reasoning process of a trading agent.
% It only requires the observable records needed to reconstruct the trading trajectory and the deployment cost.

% \stitle{Output.}
% The outputs are a standardized audit report containing profit attribution, cost attribution, viability labels, and diagnosis. Profit attribution explains whether observed profit comes from passive market exposure, initial asset selection, or dynamic reallocation. Cost attribution explains whether deployment cost is dominated by LLM inference, trading execution, infrastructure, or residual costs. The viability labels distinguish whether the whole system is profitable after all-in costs and whether the agentic intervention itself is economically justified.

\stitle{Outputs.}
\sys produces two user-friendly reports (see \add{format in} Appendix~\ref{app:format}).
The financial report summarizes portfolio performance, profit attribution, cost attribution, and viability status.
The agentic system diagnosis report explains the failure mode and provides evidence-grounded revision suggestions for the trader.

% \subsection{Measurement Layer}
\subsection{Accounting Layer}
As shown in Figure~\ref{fig:framework}, the accounting layer traces the trading trajectory and applies the attribution method defined in Section~\ref{sec:method} to profit and cost.
The profit module replays trading actions to recover portfolio values and decomposes gross profit into market, selection, and timing effects.
The cost module accounts for commissions, LLM usage, infrastructure, data subscriptions, and uncertainty costs.
Eventually, it delivers auditable results, including the action ledger, attribution tables, and diagnostic figures, providing fixed evidence for the next layer.

\subsection{Diagnosis Layer}

The diagnosis layer converts accounting results into interpretable system-level diagnoses. It checks viability, identifies dominant failure modes, and generates diagnostic reports.
% As shown in Figure~\ref{fig:framework}, 
% The diagnosis agent acts as an analyst grounded in the current system configuration, measured profit--cost components, and runtime traces.
It first extracts key metrics from the audit results and summarizes runtime records into a compact execution profile.
These structured inputs are then combined under a constrained output contract, requiring the final report to follow a fixed diagnostic structure and to ground conclusions in supplied evidence.

The diagnosis produces two types of analysis.
First, it interprets viability by combining trading outcomes and runtime measurements, identifying whether weak performance stems from insufficient trading profit, high model-side cost, excessive decisions, or execution frictions.
Second, it maps trace-supported failure modes to actionable revisions.
For example, frequent decisions with limited incremental profit indicate a cadence--return mismatch.
The agent then prioritizes operational fixes, such as reducing redundant model calls or decision frequency, and structural revisions, such as improving execution logic, portfolio construction, trading architecture, or the LLM's role in the decision loop.

% The diagnosis produces two types of analysis.
% First, it interprets viability by combining trading outcomes and runtime measurements, diagnosing whether weak performance is mainly associated with insufficient trading profit, high model-side cost, overly frequent decisions, or high execution frictions.
% Second, it maps trace-supported failure modes to actionable revisions.
% For example, frequent decisions with limited incremental profit indicate a cadence--return mismatch.
% The agent then prioritizes short-term operational fixes, such as reducing redundant model calls or decision frequency, and longer-term structural revisions, such as improving execution logic, portfolio construction, trading architecture, or the role of the LLM in the decision loop.

% This design helps practitioners understand not only whether the system is economically viable, but also why it succeeds or fails and which components should be revised.

% It first extracts key contextual variables and metrics from the financial report, including return, cost decomposition, net outcome, opportunity cost, latency, and token intensity.
% It also summarizes experiment records into a compact runtime profile, covering decision distributions, trade counts, slippage, tool use, average latency, and token consumption.
% These structured inputs are injected into a constrained output contract, allowing the agent to reason over report evidence, runtime traces, and static configurations before producing the final diagnosis.

\section{Experiments}
\label{sec:exp}

We empirically evaluate the economic viability of agentic trading systems through cost-profit breakdowns and conduct experiments on four deployment dimensions: backbone model, capital scale, trading frequency, and system architecture.
We address the following research questions (RQs):

% \begin{itemize}[leftmargin=*]
%     \item \add{\textbf{RQ1 (Backbone model)}: How do LLM backbones differ in profit sources, cost burdens, and net margin under the same trading pipeline?}
%     \item \add{\textbf{RQ2 (Capital scale)}: Can larger capital dilute fixed costs, or amplify backbone-specific trading outcomes?}
%     \item \add{\textbf{RQ3 (Trading frequency)}: Does higher decision frequency create enough marginal profit to cover additional inference and execution costs?}
%     \item \add{\textbf{RQ4 (System architecture)}: Does architectural complexity translate into decision-attributed gains after coordination and trading costs?}
%     \item \add{\textbf{RQ5 (Forward diagnosis and deployability)}: Can forward diagnosis identify deployment bottlenecks, and do practitioners find profit--cost explanations useful for revising agentic trading systems?}
% \end{itemize}

\begin{itemize}[leftmargin=*]
    \item \textbf{RQ1 (Backbone model)}: How do LLM backbones differ in profit sources, intelligence cost, and net margin under the same trading pipeline?
    \item \textbf{RQ2 (Capital scale)}: Does capital scaling improve agentic viability by diluting fixed costs, or does it amplify the value and risk of LLM-mediated decisions?
    \item \textbf{RQ3 (Trading frequency)}: Does higher decision frequency create enough marginal profit to cover additional inference and execution costs?
    \item \textbf{RQ4 (System architecture)}: Does architectural complexity translate into decision-attributed gains, or merely increase coordination and execution costs?
\end{itemize}

\subsection{Setup}
\label{sec:exp-setup}

\stitle{Deployment setting.}
We instantiate \sys on top of a representative agentic trading system, AI-Trader~\cite{fan2025aitrader}, to evaluate whether the toolkit can audit realistic decision pipelines.
The system trades a fixed universe of liquid U.S. equities over a two-month market window from December 1, 2025 to January 30, 2026, starting with an initial capital of \$100{,}000.
Market data are accessed through licensed provider APIs/subscriptions, and raw provider data are not redistributed.

\stitle{Profit--cost setting.}
We adopt a simplified but end-to-end cost formulation that covers four deployment-relevant components: LLM-side decision cost, trading-side execution cost, infrastructure cost, and stochastic runtime overhead.
On the profit side, we construct two attribution baselines: a systematic exposure baseline based on the S\&P 500 return over the same window, \add{applied only to the portfolio's initially invested capital,} and a static allocation baseline that holds the initial portfolio weights unchanged.

% Detailed cost assumptions and provider-specific pricing are reported in Appendix~\ref{app:exp-setup}.
Detailed settings are reported in Appendix~\ref{app:exp-setup}.

\stitle{Experiment scenarios.}
For backbone model (RQ1), we compare flagship LLMs from leading providers, including 
GPT-5.2~\cite{openai2025gpt52}, 
Gemini 3 Flash~\cite{google2025gemini3flash}, 
Claude Sonnet 4.5~\cite{anthropic2025claude}, 
Qwen3-Max~\cite{yang2025qwen3}, 
DeepSeek-V3.2~\cite{deepseekai2025deepseekv32}, 
GLM-4.7~\cite{zai2025glm47}, 
Kimi-K2~\cite{team2025kimi}, 
MiniMax-M2.1~\cite{MiniMaxM21_2025}, 
Llama-4-scout~\cite{MetaLlama4_2024}, 
and Mistral-Large-3~\cite{mistral2025mistrallarge3}. 
For RQ2--RQ4, we utilize GPT-5.2 and DeepSeek-V3.2 as representative backbones. 
RQ2 varies the initial endowment from \$10{,}000 to \$500{,}000 and reports net profit, cost-to-capital ratio, and break-even status. RQ3 compares daily and hourly decision frequencies using marginal profit minus marginal cost in one month horizon. 
% RQ4 compares architectures with adjusting reasoning and coordination complexity. We utilize CoT, tool-augmented, and multi-agent configurations.
RQ4 explores the architecture reasoning complexity by placing AI-Trader as a moderate design. We utilize the typical Chain-of-Thought prompting (See details in Appendix~\ref{app:rq4}) and DeepFund~\cite{li2025time} as simple and complex designs, respectively.
All variants use the same trading logic and execution assumptions.

\subsection{Results Analysis (RQ1--RQ4)} 
% \subsection{Offline Experiments (O1--O4)}
\label{sec:exp-o1-o4}

% The observations are configuration- and window-specific and are not intended to establish intrinsic trading ability or universally generalizable model rankings.
Our analysis focuses on viability patterns across the evaluated configurations and market windows.

\begin{figure}
    \centering
    \includegraphics[width=\linewidth]{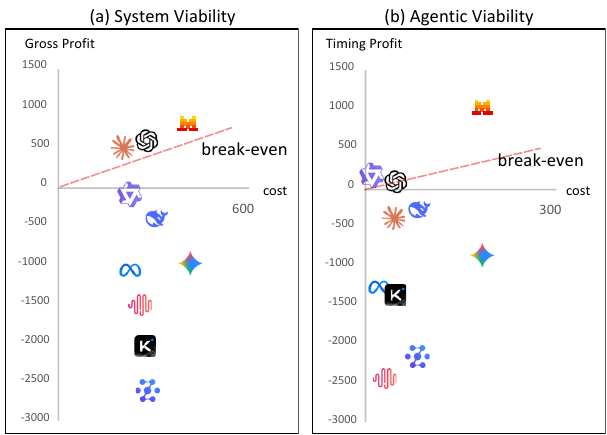}
    \caption{\textbf{Viability across backbone models.}
    (a) System viability and (b) agentic viability:
    profit (y-axis) versus cost (x-axis) for 10 backbone LLMs under identical trading configurations.}
    \label{fig:q1-backbone}
\end{figure}

\begin{table*}[t]
\centering
% \small
\resizebox{\linewidth}{!}{%
\begin{tabular}{lrrrrrrrrrr}
\toprule
\textbf{Metric}                                                           & \textbf{DeepSeek-V3.2} & \textbf{MiniMax-M2.1} & \textbf{Kimi-K2} & \add{\textbf{GPT-5.2}} & \textbf{GLM-4.7} & \textbf{Llama-4-scout} & \add{\textbf{Qwen3-Max}} & \textbf{Claude Sonnet 4.5} & \textbf{Gemini 3 Flash} & \textbf{Mistral-Large-3} \\ \midrule
\textbf{Gross Profit \textsuperscript{(a)}}              & -388.29                & -1474.62              & -1975.63         & 505.76             & -2554.81         & -1020.38               & -50.43           & 519.40                     & -975.58                 & 777.81                   \\
\quad Market Effect                                        & 1351.45                & 1718.95               & 1714.40          & 1535.80            & 1696.78          & 1521.73                & 0.00             & 1716.01                    & 1351.45                 & 1719.04                  \\
\quad Asset Selection                                         & -1500.44               & -813.13               & -2399.00         & -1125.69           & -2099.53         & -1304.89               & -175.90          & -870.58                    & -1500.44                & -1981.37                 \\
\quad Timing Effect \textsuperscript{(1)} & -239.30                & -2380.44              & -1291.03         & 95.65              & -2152.06         & -1237.22               & 125.46           & -326.02                    & -826.59                 & 1040.14                  \\ \midrule
\textbf{Total Cost \textsuperscript{(b)}}                & 304.94                 & 252.80                & 275.67           & 258.79             & 296.95           & 228.90                 & 223.04           & 255.04                     & 422.38                  & 414.82                   \\
\quad Trading \textsuperscript{(2)}       & 63.00                  & 33.00                 & 56.00            & 32.00              & 79.38            & 10.00                  & 4.00             & 23.00                      & 200.00                  & 195.00                   \\
\quad LLM \textsuperscript{(3)}           & 23.84                  & 1.35                  & 2.06             & 9.03               & 0.83             & 0.04                   & 0.90             & 11.38                      & 4.34                    & 1.50                     \\
\quad Infra                                                & 208.20                 & 208.20                & 208.20           & 208.20             & 208.20           & 208.20                 & 208.2            & 208.20                     & 208.20                  & 208.20                   \\
\quad Stochastic \textsuperscript{(4)}    & 9.90                   & 10.25                 & 9.42             & 9.56               & 8.55             & 10.66                  & 9.93             & 12.47                      & 9.84                    & 10.11                    \\ \midrule
\textbf{Net Profit \textsuperscript{(a-b)}}              & -693.23                & -1727.42              & -2251.30         & 246.97             & -2851.76         & -1249.29               & -273.47          & 264.36                     & -1397.96                & 362.99                   \\ \midrule
\textbf{Agentic Profit \textsuperscript{(1-2-3-4)}}      & -336.04                & -2425.04              & -1358.50         & 45.06              & -2240.81         & -1257.92               & 110.63           & -372.86                    & -1040.76                & 833.52                   \\ \bottomrule
\end{tabular}
}
\caption{Viability across backbone models. Net profit measures system viability, while agentic profit measures whether timing value covers decision-induced costs.
\add{Market Effect applies only to initially invested capital; a zero value indicates that all capital was initially held in cash.}
}

% \vspace{-1.2em}
\label{tab:rq1-model}
\end{table*}

\stitle{RQ1 (Backbone model).}
The key difference across LLM backbones lies less in their operating cost than in how they generate, or fail to generate, active profit.
Figure~\ref{fig:q1-backbone} and Table~\ref{tab:rq1-model} reveal models viability levels to four groups.
% First, Mistral-Large-3 represents a fully viable backbone: it achieves positive net profit and positive agentic profit, despite relatively high total cost. Its viability is mainly supported by a large positive timing effect, indicating that strong active reallocation can offset operating burden.

First, GPT-5.2 and Mistral-Large-3 are viable at both system and agentic levels. Mistral-Large-3 achieves the strongest agentic profit despite relatively high cost, with a large positive timing effect suggesting dynamic reallocation as a candidate driver relative to the static baseline.

Second, Claude Sonnet 4.5 represents a system-viable but agentically weak backbone.
% It achieves positive net profit, but its agentic profit is negative, suggesting that the system can cover costs while the dynamic agentic component fails to add value.
% This case shows why system viability and agentic viability should be evaluated separately.
\add{
It achieves positive net profit but negative agentic profit, indicating that the system covers costs while the dynamic component adds no incremental value. This case motivates evaluating system and agentic viability separately.}
%
% Third, most remaining backbones are non-viable, but for different reasons.
% Some models, such as DeepSeek-V3.2, suffer mainly from poor asset selection; others, such as GLM-4.7 and MiniMax-M2.1, are dominated by negative timing effects.
% Qwen3-Max is a borderline case with positive agentic profit but negative net profit, indicating that active improvement is insufficient to overcome total system cost.

Third, Qwen3-Max is agentically viable but system-non-viable: its positive agentic profit is insufficient to offset total system cost.

Fourth, the remaining six backbones are non-viable at both levels in the observed window. DeepSeek-V3.2 is dominated by negative selection, while GLM-4.7 and MiniMax-M2.1 show larger negative timing effects.

Overall, backbone choice changes not only the magnitude of profit and cost, but also the source of economic value and failure.

% % \begin{tcolorbox}[colback=yy!30!white, colframe=rr]
% \begin{tcolorbox}[
%     colback=yy!30!white,
%     colframe=rr,
%     before skip=4pt,
%     after skip=4pt,
%     left=1mm,
%     right=1mm,
%     top=1mm,
%     bottom=1mm
% ]
% \textbf{Takeaway}: 
% Backbones differ mainly in how similar intelligence costs are converted into timing value and net margin.
% % Under the evaluated setting, backbones exhibit distinct viability profiles, with differences arising primarily in decision-attributed value rather than operating cost alone.
% \end{tcolorbox}

% \begin{tcolorbox}[colback=yy!30!white, colframe=rr, 
% before skip=4pt,
% after skip=4pt,]

% \textbf{Takeaway}: 
% Backbones differ mainly in how similar intelligence costs are converted into timing value and net margin.

\begin{figure}[t]
    \centering
    \includegraphics[width=\linewidth]{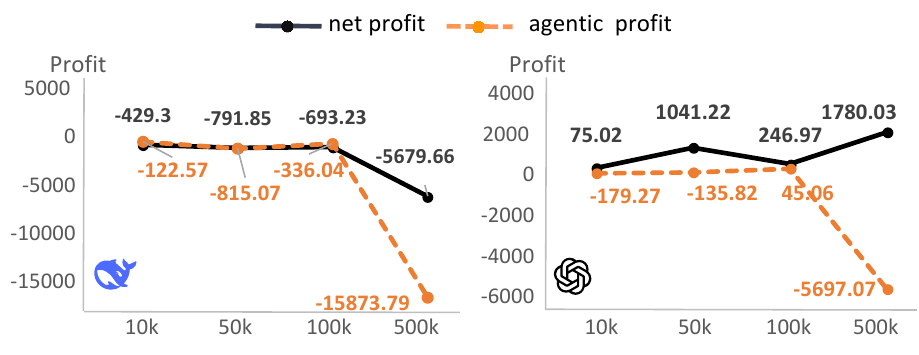}
    % \caption{\textbf{Viability across capital scale.} Cumulative system cost and net profit under varying initial cash reserves for GPT-5.2 and DeepSeek-V3.2.}
    \caption{\textbf{Viability across capital scale.} \emph{Net Profit} and \emph{Agentic Profit} under varying initial cash investment scale for GPT-5.2 and DeepSeek-V3.2.}
    \label{fig:q2-capital}
\end{figure}

\stitle{RQ2 (Capital scale).}
% \notice{[REWRITE] delete optimizer; can larger capital dilute costs? add metrics(should be predefined)}
% Figure~\ref{fig:q2-capital} shows that cumulative system cost remains relatively stable as capital scale increases for both backbones.
% In contrast, net profit exhibits pronounced scale-dependent behavior.
% For GPT-5.2, net profit increases monotonically with capital scale and transitions from marginal loss to clear profitability at larger capital levels, revealing a well-defined break-even threshold.
% DeepSeek-V3.2, however, demonstrates substantially higher volatility: while losses are moderate at small and mid-scale capital, net profit deteriorates sharply at larger scales, resulting in significant drawdowns despite comparable system costs.
% This divergence indicates that capital scaling interacts strongly with backbone-specific profit dynamics.
% While increased capital can amplify profitable strategies, it can also magnify losses, suggesting that capital scaling alone is insufficient to guarantee economic viability.
%
Figure~\ref{fig:q2-capital} shows that capital scaling does not uniformly improve viability.
DeepSeek-V3.2 exhibits a negative pattern. 
Both system and agentic viability remain negative across all scales, with losses expanding sharply at 500k. 
These losses are mainly associated with unfavorable timing effects rather than operating costs. 
% \sys identifies a gap between positive asset selection and poor execution, 
% where heavy LLM usage and timing slippage prevent the system from capturing market upside.
%
% For GPT-5.2, system viability remains positive and generally increases with capital, whereas agentic viability is non-linear. 
For GPT-5.2, system viability remains positive but varies non-monotonically with capital scale, whereas agentic viability is also non-linear.
However, with 500k initial cash, it only gains a modest return compared to the setting with 50k, and this result is mainly due to passive market exposure, while the timing effect \add{remains negative.}
The diagnosis therefore recommends streamlining the inference and order pipeline to reduce latency and slippage, and recover timing value.

Overall, capital scale acts as more than a simple cost-dilution mechanism, as it may amplify both favorable timing outcomes and strategy-level losses.
See Appendix~\ref{app:rq2} for more details.

% % \begin{tcolorbox}[colback=yy!30!white, colframe=rr]
% \begin{tcolorbox}[
%     colback=yy!30!white,
%     colframe=rr,
%     before skip=4pt,
%     after skip=4pt,
%     left=1mm,
%     right=1mm,
%     top=1mm,
%     bottom=1mm
% ]
% \textbf{Takeaway.} Capital scaling does not simply dilute fixed costs; it amplifies the value and risk of model's timing behavior.
% % Capital scaling reveals whether viability changes arise from fixed-cost dilution or from changes in decision-attributed value.
% \end{tcolorbox}

\begin{figure}[t]
    \centering
    \includegraphics[width=\linewidth]{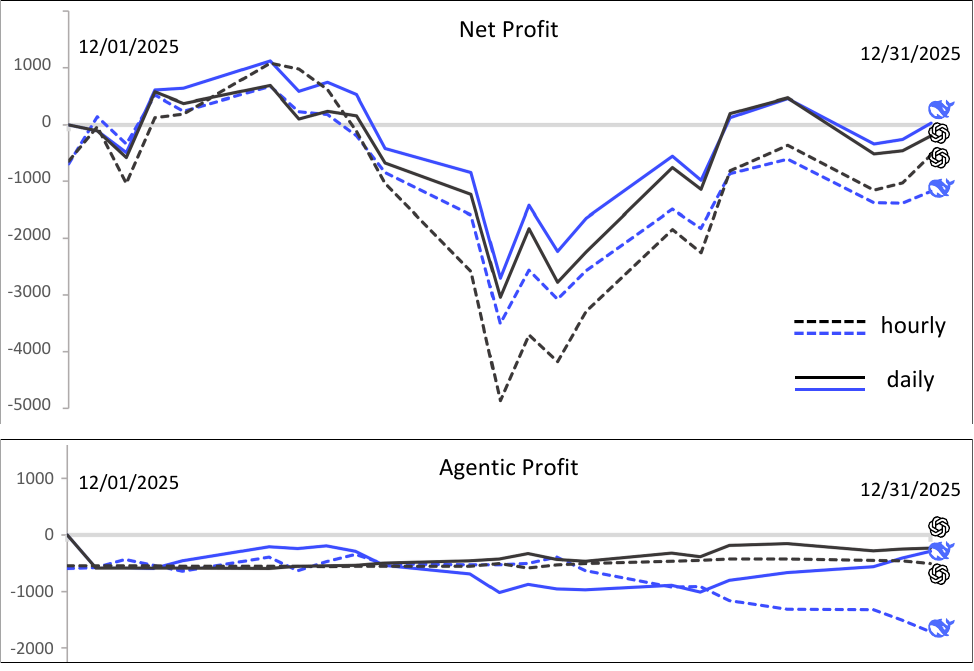}
    \caption{\textbf{Viability across trading frequency.} Cumulative net profit and agentic profit over time under hourly (dashed) and daily (solid) trading frequencies for GPT-5.2 and DeepSeek-V3.2. The horizontal line at zero indicates the break-even line.}
    \label{fig:q3-frequency}
\end{figure}

\stitle{RQ3 (Trading frequency).}
% Figure~\ref{fig:q3-frequency} compares economic outcomes under hourly and daily decision frequencies.
% For GPT-5.2, increasing the decision frequency from daily to hourly raises cumulative system cost from about \$141 to approximately \$163, reflecting higher inference usage and execution overhead.
% For DeepSeek-V3.2, the cost impact is more pronounced: hourly trading nearly doubles LLM-side inference cost (around \$311) relative to the daily setting (around \$163), indicating strong cost sensitivity to decision frequency.
% This suggests that higher decision frequency scales inference costs and execution frictions at a rate that outpaces the marginal improvement in gross returns. 
% Although higher-frequency trading partially improves performance for both models with more days of positive net profit, they still fail to cover the total costs and remain below the break-even line. These results suggest that increases in trading frequency should be applied with caution and only under market conditions where higher-frequency decisions can reliably translate into positive marginal returns.
%
Figure~\ref{fig:q3-frequency} compares cumulative net profit and agentic profit under hourly and daily decision frequencies. 
The results show that higher decision frequency does not improve viability. 
For both backbones, daily trading outperforms hourly trading in terms of both net profit and agentic profit.
DeepSeek-V3.2 improves from a net profit of -1222.24 under hourly trading to 29.84 under daily trading, while GPT-5.2 improves from -585.46 to -198.28.
% This pattern is not only caused by higher operating costs.
%
% The diagnosis shows that the hourly setting changes the quality of active decisions.
% \add{The diagnosis shows that the hourly setting is associated with less favorable active-decision outcomes.}
This stems not merely from higher operating costs, but from degraded active-decision quality under the hourly setting.

Hourly mode raises total cost for both and might produce worse gross profit, especially for DeepSeek-V3.2~(see Appendix Table~\ref{tab:rq3-fre}).
This suggests that more frequent decisions introduce additional trading noise and timing errors.

%  % \begin{tcolorbox}[colback=yy!30!white, colframe=rr]
% \begin{tcolorbox}[
%     colback=yy!30!white,
%     colframe=rr,
%     before skip=4pt,
%     after skip=4pt,
%     left=1mm,
%     right=1mm,
%     top=1mm,
%     bottom=1mm
% ]
% \textbf{Takeaway.} Higher frequency fails when extra decisions add noisy trades and timing errors.
% % In the evaluated configurations, higher decision frequency is associated with worse viability primarily through degraded decision-attributed value rather than operating cost alone.
% \end{tcolorbox}

% \textbf{Takeaway.} Higher frequency fails when extra decisions add noisy trades and timing errors.

\begin{table*}[t] 
\centering
\small
\resizebox{0.8\linewidth}{!}{%
\begin{tabular}{l ccc c ccc}
\toprule
& \multicolumn{3}{c}{\textbf{DeepSeek-V3.2}} & & \multicolumn{3}{c}{\textbf{GPT-5.2}} \\ 
\cmidrule(lr){2-4} \cmidrule(lr){6-8}
\textbf{Metric} & \textbf{CoT} & \textbf{AI-Trader} & \textbf{DeepFund} & & \textbf{CoT} & \textbf{AI-Trader} & \textbf{DeepFund} \\ 
\midrule
\textbf{Gross Profit \textsuperscript{(a)}} & -1236.25 & -388.29 & 610.00 & & 341.87 & 505.76 & 881.25 \\
\quad Market Effect & 514.70 & 1351.45 & 150.25 & & 1432.69 & 1535.80 & 1338.89 \\
\quad Asset Selection & 1463.8 & -1500.44 & -56.88 & & -518.61 & -1125.69 & -902.11 \\
\quad Timing Effect \textsuperscript{(1)} & -3214.75 & -239.30 & 516.62 & & -572.20 & 95.65 & 444.47 \\
\midrule
\textbf{Total Cost \textsuperscript{(b)}} & \add{278.57} & 304.94 & \add{397.02} & & \add{245.44} & 258.79 & \add{403.11} \\
\quad Trading \textsuperscript{(2)} & 60.00 & 63.00 & 174.00 & & 25.23 & 32.00 & 178.00 \\
\quad LLM \textsuperscript{(3)} & 1.35 & 23.84 & 3.94 & & 1.27 & 9.03 & 6.13 \\
\quad Infra & 208.20 & 208.20 & 208.20 & & 208.20 & 208.20 & 208.20 \\
\quad Stochastic \textsuperscript{(4)} & 9.02 & 9.90 & 10.67 & & 10.75 & 9.56 & 10.59 \\
\midrule
\textbf{Net Profit \textsuperscript{(a-b)}} & -1514.82 & -693.23 & 212.98 & & 96.43 & 246.97 & 478.14 \\
\midrule
\textbf{Agentic Profit \textsuperscript{(1-2-3-4)}} & -3285.12 & -336.04 & 328.01 & & -609.45 & 45.06 & 249.76 \\
\bottomrule
\end{tabular}
}
\caption{Viability across system architecture with \$100{,}000 initial cash.}
\label{tab:rq4-system}
\end{table*}

\stitle{RQ4 (System architecture).}
% Table~\ref{tab:complexity} compares economic outcomes across varying complexity tiers by analyzing the shift from CoT to Multi-agent configurations. In the case of DeepSeek-V3.2, increased complexity (from Zero-shot to Multi-agent) leads to higher gross profits, but the additional costs bring the system into profitability only in the multi-agent configuration. For GPT-5.2, however, adding complexity seems to reduce net profitability, as the coordination and reasoning costs outweigh the benefits of higher system complexity. This suggests that system architecture should be carefully balanced with the economic viability of the model, as increasing complexity can lead to diminishing returns if the additional costs do not result in proportionate gains. 
%
%
Table~\ref{tab:rq4-system} shows that architectural design affects viability through active decision quality rather than cost alone.
CoT has the lowest total cost for both backbones, but it does not produce the best outcomes, suggesting that low-cost reasoning alone is insufficient for trading.
The Appendix~\ref{app:rq4} diagnosis further shows that this failure is not driven by cost accumulation 
% and the key bottleneck is when the agent reallocates capital.
\add{and the largest negative contribution appears in the dynamic reallocation.}
By contrast, DeepFund achieves the strongest performance: it is the only architecture with positive net profit and agentic profit for DeepSeek-V3.2, and it also yields the highest values for GPT-5.2.
Despite its higher cost,
% DeepFund's advantage comes mainly from timing.
\add{DeepFund's stronger observed performance mainly comes from its positive timing.}
% It produces positive timing effects for both backbones, showing that its additional reasoning and coordination are converted into better dynamic reallocation.
\add{It yields positive timing effects for both backbones, indicating more favorable dynamic reallocation under the evaluated setting.}
Thus, the value of an agentic architecture depends not on being cheaper or more complex, but on whether it can convert reasoning cost into better asset selection and timing decisions.

We further report complementary financial metrics (see Appendix~\ref{app:financial_metrics}) to characterize the realized return and risk of the evaluated portfolios, together with an extended six-month evaluation using representative frontier backbones (see Appendix~\ref{app:extended-exp}).

\section{Discussion}
% \stitle{}?
\stitle{Sensitivity in Market Regime.}
Because trading performance is highly regime-dependent, we further examine viability and find that market regimes reshape profit sources more strongly than system costs, making attribution necessary for interpreting deployability across environments.
Detailed results are provided in Appendix~\ref{app:regime}.

% \stitle{Forward Diagnosis and Practitioner Feedback.}
% We further evaluate \sys from a deployment-oriented perspective. 
% We conduct a forward diagnostic study that uses findings from a December window to inform a January 2026 deployment comparison, and we collect feedback from 13 retail traders and 5 external reviewers. 
% These analyses examine whether profit-cost diagnosis can help users interpret non-viability, distinguish cost-driven and strategy-driven failure modes, and identify revision directions for deployment. 
% Detailed results and examples are provided in Appendix~\ref{deployment}.

\stitle{Forward Diagnosis.}
We test whether \sys provides useful signals for later deployment decisions.
Using the December diagnostic window, it flags hourly trading as cost- and friction-intensive, motivating a January 2026 daily-versus-hourly comparison.
Although daily trading performs better in early January, both settings remain net negative, showing that cost control alone cannot ensure viability when strategy-level profit generation is weak.
Detailed results are in Appendix~\ref{app:forward}.

\stitle{Practitioner Feedback.}
We also assess whether the diagnosis is interpretable and useful in private deployments.
We collect feedback from 13 retail traders deploying \sys.
Participants find the joint profit--cost diagnosis useful for distinguishing weak decision value, trading friction, and LLM usage.
They also emphasize that actionable deployment support should go further by providing stronger strategy-level optimization guidance.
Additional details are in Appendix~\ref{app:deploy}.

\section{Conclusion and Future Work}
\label{sec:conclusion}

We introduced \sys, a trace-grounded profit--cost viability diagnosis toolkit for evaluating agentic trading systems.
It connects trading outcomes, runtime traces, and deployment costs under a shared accounting window, allowing users to assess whether a system is profitable as a whole and whether its LLM-mediated decisions generate enough incremental value to justify the induced costs.
By distinguishing system from agentic viability, \sys diagnoses whether intelligence is converted into economic value.
Future work will extend \sys to more asset classes, refine execution friction modeling, and optimize agent architectures toward sustainable economic returns.

% Future work can improve friction estimation, extend the toolkit to additional asset classes and execution venues, and study how tool use, multi-agent coordination, and long-context reasoning affect the viability frontier.
% Viability signals may also be integrated with reinforcement learning~\cite{jin2025controlling,li2024towards,qian2025xrouter} to optimize agentic systems toward better economic outcomes.

% Overall, our findings reframe the evaluation of LLM-based trading agents from capability-centric performance ranking to trace-grounded diagnosis of whether agentic intelligence can pay for itself.

\clearpage
% \newpage

\section*{Limitations}
% \qi{done, pls check}

% This paper focuses on evaluating agentic trading systems themselves, not on proposing a general trading strategy.
This work uses controlled experimental settings to demonstrate how \sys reveals profit--cost behaviors in agentic trading systems.
% Our experiments use data collected from an open-source agentic system and are limited to 10 NASDAQ stocks over short evaluation windows.
%
Thus, the reported patterns should be viewed as diagnostic observations under specific settings, not general claims about the intrinsic trading ability of any backbone model.
Their generalizability to other asset classes, broader investment universes, longer horizons, or different market regimes remains to be validated.
We use the S\&P 500 as a broad-market benchmark proxy in our experiments. 
This choice may introduce slight attribution bias in measuring systematic risk exposure, and users can customize the benchmark to better match their own trading universe and deployment needs.
The practitioner evaluation involves a modest number of participants (13 retail traders) and collects qualitative feedback rather than controlled experimental evidence of downstream decision improvement.
Additionally, cost inputs such as infrastructure fees and stochastic terms are user-configured rather than automatically measured; accuracy, therefore, depends on the fidelity of the values provided by the user.

\section*{Future Work}
Future work can improve friction estimation, extend the toolkit to more asset classes and execution venues, and study how tool use, multi-agent coordination, and long-horizon reasoning affect the viability frontier.
At the architecture level, controlled ablations could assess whether additional tool calls, specialized agents, or longer historical contexts provide incremental decision value sufficient to offset their computational and coordination costs.
Such analyses could identify when additional reasoning is economically justified and when a simpler configuration offers comparable outcomes at lower cost.

We also plan to evaluate broader market regimes, incorporate finer-grained latency-to-fill logs and slippage models, and compare against stronger non-LLM baselines. For instance, evaluations spanning trending, sideways, and high-volatility markets could test whether observed profit sources remain stable as market conditions change.
Notably, viability signals could be integrated with reinforcement learning~\cite{jin2025controlling,li2024towards} and human preference~\cite{ong2025routellm} to optimize agentic systems toward better economic outcomes. 

For example, training objectives could incorporate deployment costs and trading friction alongside returns and risk, while practitioner preferences could guide trade-offs among profitability, interpretability, and operational complexity.

Finally, these extensions strengthen \sys as a framework for diagnosing and improving the economic viability of agentic trading systems.

\section*{Ethical Considerations}

The present work has been conducted solely for academic research purposes and does not constitute investment advice or a recommendation to process any financial asset.
Users deploying agentic trading systems in real markets should exercise judgment and comply with applicable regulations. We acknowledge that cost-aware evaluation, while promoting transparency, does not eliminate the risks inherent in algorithmic trading, including potential market impact, amplification of biases in LLM reasoning, and the possibility that cost-optimized configurations may inadvertently encourage excessive risk-taking.

All market data used in our experiments is publicly available third-party data accessed through licensed provider APIs, and we do not redistribute any raw data.
The practitioner feedback study involved voluntary participants, and we collected only qualitative feedback without personally identifiable information, sensitive information, private trading strategies, or portfolio details.
The diagnostic reports generated by \sys are intended as informational analysis tools and should not be interpreted as financial advice. 
%

% Acknowledgments go here (omit in review version).

\section*{Acknowledgments}
This paper was supported by the NSF of China (62402409); Youth S\&T Talent Support Programme of Guangdong Provincial Association for Science and Technology (SKXRC2025461); Guangdong provincial project (Project No. 2023CX10X008); 
Young Talent Support Project of Guangzhou Association for Science and Technology (QT-2025-001); Guangzhou Science and Technology Program (2026C99J0012); Guangzhou Basic and Applied Basic Research Foundation (2026A1515010269, 2025A04J3935, 2023A1515110545); and Guangzhou-HKUST(GZ) Joint Funding Program (2025A03J3714). Lastly, we genuinely thank the reviewers for their constructive suggestions during the rebuttal period.

% Bibliography (style is set automatically by acl.sty to acl_natbib)
\bibliography{main}
% \newpage

\appendix

\newpage 
\section{Diagnosis Report Format}
\label{app:format}

In this part, we provide two formats to show what \sys outputs in practice; a complete output example is provided in our open source repository. 
First, we present an example of a financial report, which reports the exact profit and loss and other trading performance metrics of a trial run. 
Based on a detailed financial analysis report, \sys generates the diagnosis report with ranked suggestions from multiple dimensions, including capital budget, backbone model choice, trading frequency, and architectural intelligence such as modules, agents, and tools.\\

\begin{tcolorbox}[
title={Financial Report Format},
nobeforeafter,
breakable,
left=1mm,right=1mm,top=1mm,bottom=1mm,colframe=teal!50,colback=white]
\begin{enumerate}
    \item Trading Configuration
        \begin{itemize}
            \item Trading Period: 
            \item Trading Model: 
            \item Assets: 
        \end{itemize}
        
    \item Asset \& Portfolio State
        \begin{itemize}
            \item Cash condition: 
            \item Positions:
        \end{itemize}
        
    \item Performance
        \begin{itemize}
            \item Profit Attribution and figure
            \item Cost Breakdown and figure
            \item Financial Statement         
        \end{itemize}
        
    \item Execution Quality
        \begin{itemize}
            \item Opportunity Cost and latency
            \item Execution layer
        \end{itemize}
\end{enumerate}

\end{tcolorbox}

\newpage
\begin{tcolorbox}[
title={Diagnosis Report Format},
nobeforeafter,
breakable,
left=1mm,right=1mm,top=1mm,bottom=1mm,colframe=teal!50,colback=white]

\textbf{Executive Summary} \\
Overview of the overall report and top suggestions

\begin{enumerate}
    \item Trading Performance Analysis
        \begin{itemize}
            \item Outcome Snapshot.
            \item Profit Attribution Analysis.
            \item Cost Structure Analysis.
            \item Portfolio Timeline and Risk Signals.
            \item Execution Quality and LLM Efficiency.    
        \end{itemize}
        
    \item Trading System Diagnosis
        \begin{itemize}
            \item System Diagnosis. \\
            Define core restrictive factors and auxiliary limiting conditions of strategy performance; assess the impact of fee expenditure on actual revenue.
            
            \item Root Causes. \\
            Summarize fundamental issues behind unsatisfactory performance based on analytical data.
            
            \item Top Recommended Actions. \\
            Provide recommendations targeting model parameters, capital allocation, and system framework; predict improvement effects and outline specific implementation approaches.
            
            \item Quick Wins vs. Structural Changes. \\
            Short-term adjustable optimization measures vs. long-term systematic reconstruction and upgrade plans.
        \end{itemize}
\end{enumerate}

\end{tcolorbox}
\newpage
\section{Profit Attribution Details}
\label{app:profit}

% \qi{need to rewrite, in the same structure with cost}
As presented in Table~\ref{tab:profit-categories}, we attribute cumulative gross profit to three portfolio-level components. The decomposition is defined by two counterfactual baselines constructed on the same asset universe and evaluation window.

\begin{table}[htbp]
    
    \small
    \begin{center}
        \begin{tabular}{p{0.18\linewidth} p{0.22\linewidth} p{0.48\linewidth}}
            \toprule
            \textbf{Component} & \textbf{Counterfactual} & \textbf{Economic interpretation} \\
            \midrule
            \textbf{Market effect} &
            Market return &
            Profit from exposure to aggregate market movement in the investable universe \\
            \midrule
            \textbf{Asset selection} &
            Static initial allocation &
            Gain from the chosen initial weights relative to market exposure \\
            \midrule
            \textbf{Timing} &
            Dynamic reallocation &
            Incremental gain from changing positions after the initial allocation \\
            \bottomrule
        \end{tabular}
    \end{center}
    \caption{Profit attribution categories for agentic trading systems.}
    \label{tab:profit-categories}
\end{table}

\stitle{Notation.}
Let \(t\in\{1,\ldots,T\}\) index decision periods, \(N\) the number of tradable assets, \(w_i^{(t)}\) the portfolio weight of asset \(i\) at period \(t\) with \(\sum_i w_i^{(t)}\leq1\), and \(r_i^{(t)}\) its realized return. Let \(V_0\) denote the initial portfolio value, \(\alpha_0=\sum_i w_i^{(0)}\) the initially invested fraction, and \(\tau_0\) the first period in which the agent establishes a non-cash allocation.

\stitle{Systematic exposure baseline.}
Let \(r_t^{\mathrm{mkt}}\) denote the realized market return in period \(t\), reflecting aggregate market movement in the investable universe. In \sys, \(r_t^{\mathrm{mkt}}\) is computed from the benchmark market prices supplied in the audit inputs. The counterfactual value under passive market exposure is

$$
    V_T^{\mathrm{sys}}
    = V_0\left[(1-\alpha_0)+\alpha_0\prod_{t=1}^{T}\left(1+r_t^{\mathrm{mkt}}\right)\right].
$$

The attributed profit is \(P^{\mathrm{sys}}=V_T^{\mathrm{sys}}-V_0\). Thus, an agent starting entirely in cash has zero Market Effect.

\stitle{Static allocation baseline.}
The static baseline holds the agent's first active allocation, established at \(\tau_0\), unchanged for the remainder of the evaluation period:

$$
    V_T^{\mathrm{base}}
    = V_0 \prod_{t=\tau_0}^{T}\left(1+\sum_{i=1}^{N} w_i^{(\tau_0)} r_i^{(t)}\right).
$$

The attributed profit is \(P^{\mathrm{asset}}=V_T^{\mathrm{base}}-V_T^{\mathrm{sys}}\). Hence, an agent may have zero Market Effect but non-zero Asset Selection if it starts in cash and establishes its first active allocation later.

\stitle{Relation to classical attribution.}
Classical Brinson attribution compares portfolio and benchmark weights at the security level~\cite{brinson1986determinants,brinson1991determinants}. Our formulation operates at the portfolio-value level and uses two nested counterfactual paths rather than a single external benchmark. This design is tailored to agentic trading systems, where decisions are naturally described as market exposure, initial allocation, and subsequent reallocation, and where the evaluation goal is deployability diagnosing rather than performance ranking.

\section{Cost Taxonomy}
\label{app:cost}

Following Williamson's transaction cost economics~\cite{Williamson2005}, markets are not frictionless: every exchange requires resources for information processing, decision-making, and execution, beyond price movement. Such transaction costs arise from bounded rationality and  coordination actions under uncertainty, and they directly shape whether a strategy is viable in practice. This perspective is especially relevant for agentic trading systems, where ``intelligence'' is produced via paid inference and deployed through fee and impact sensitive execution pipelines. Motivated by this view, we construct an explicit cost model that covering end-to-end frictions across decision-making, trading execution, and infrastructure.

Let \(t \in \{1,\dots,T\}\) index decision periods within a time window \(T\) (e.g., minutes, hours, or days). We decompose the total cost in period \(t\) into LLM-side, trading-side, infra-side, and stochastic:
\begin{equation}
    C_t \;=\; C_t^{\text{llm}} \;+\; C_t^{\text{trd}} \;+\; C_t^{\text{inf}} \;+\; C_t^{\text{sto}}
    \label{eq:daily-cost}
\end{equation}

Specifically, the components are computed as follows.
The trading-side cost \(C_t^{\text{trd}}\) combines fixed charges and
activity-dependent expenses,
\[
C_t^{\text{trd}}
=
C_{\text{stat}}^{\text{trd}}
+
\sum_{i=1}^{M_{\text{trd}}}
\gamma_{t,i}^{\text{trd}} \cdot \text{activity}_{t,i}^{\text{trd}}.
\]

The LLM-side cost \(C_t^{\text{llm}}\) is determined by token usage and adjusted by the success rate,
\[
c_t^{\text{llm}}
=
\frac{\sum_{s \in \mathcal{S}} p^{s} \cdot x_r^{s}}{\rho},
\qquad
\mathcal{S}=\{\text{in},\,\text{out},\,\text{cache}\},
\]
where \(\rho\) is successful completion rate due to retries.

The infra-side cost is modeled as a constant per period,
\[
C_t^{\text{inf}} = \kappa,
\]
and the stochastic cost captures unanticipated expenses,
\[
C_t^{\text{sto}} \sim \text{Uniform}(0, n).
\]

Given the cost taxonomy defined above, \sys leverages a \emph{system-agnostic} cost accounting module that acts as a lightweight middleware between an agent's execution pipeline and the downstream accounting logic. Instead of assuming a specific prompting strategy or agent architecture, the taxonomy defines a small set of standardized events (with timestamps and identifiers) so that agentic workflows can all be logged and attributed consistently to a decision period and step. The dynamic portion of each category (\eg token-dependent inference fees, per-trade frictions such as commissions or slippage) is traced during execution through event logging. In contrast, the static portion (\eg data subscription or deployment baseline fees) is configured by end users as part of the evaluation setting, enabling \sys to reflect different billing plans and deployment assumptions without modifying the agent.
Concretely, it considers the following aspects:

\begin{itemize}[leftmargin=*]
    \item \textbf{LLM-side accounting} \(C_t^{\text{llm}}\): converts traced model usage (prompt/completion tokens, call counts, retries) into monetary inference cost under provider pricing (or user-specified prices).
    \item \textbf{Trading-side accounting} \(C_t^{\text{trd}}\): aggregates execution-layer costs from orders and fills, including explicit fees (commissions, exchange/clearing fees, taxes) and modeled frictions (spread/slippage) when measurable.
    \item \textbf{Infra-side accounting} \(C_t^{\text{inf}}\): aggregates deployment costs such as elastic compute/bandwidth/I/O and monitoring/logging overhead; static baselines (\eg hosting plans) are incorporated from user configuration, while dynamic usage is traced during execution.
    \item \textbf{Stochastic term} \(C_t^{\text{sto}}\): represents unanticipated or hard-to-measure expenses as an additive uncertainty term; it can be set to zero (best-case) or calibrated from real billing data (conservative-case) as discussed in previous work~\cite{nevmyvaka2006reinforcement,gatheral2010no}.
\end{itemize}

As presented in Table~\ref{tab:cost-categories}, we categorize the cost 
items to four types according to (i) where the cost arises, and (ii) whether the cost is static (fixed
for a period or billing cycle) or dynamic. 
\add{As Stochastic term is sampled at runtime, minor cost differences may occur across independent runs.}

\begin{table*}[htbp]
    \small
    \begin{center}
        \resizebox{\textwidth}{!}{%
        \begin{tabular}{p{0.1\linewidth} p{0.25\linewidth} p{0.6\linewidth}}
            \toprule
            \textbf{Category} & \textbf{Static (fixed)} & \textbf{Dynamic (variable)} \\
            \midrule
            \textbf{LLM-side} &
            Model subscription/licensing (if applicable) &
            Inference cost (prompt/completion tokens, context length, number of agent/tool calls, retries) \\
            \midrule
            \textbf{Trading-side} &
            Market data subscription; account/brokerage plan fees &
            Commissions/transaction fees; taxes (\eg stamp duty); exchange/clearing fees; bid--ask spread and slippage (market impact) \\
            \midrule
            \textbf{Infra-side} &
            Hosting/deployment baseline; databases and storage &
            Elastic infrastructure usage (compute, bandwidth, I/O); monitoring/logging and reruns \\
            \midrule
            \textbf{Stochastic} &
            Miscellaneous fixed charges not explicitly modeled &
            Residual/unanticipated costs (\eg unexpected platform fees, outages and reruns, sudden price changes) \\
            \bottomrule
        \end{tabular}%
        }
    \end{center}
    \caption{Cost categories for end-to-end accounting.}
    \label{tab:cost-categories}
\end{table*}
\section{Experiment Detail}

Across all experiments, we gathered data through an agentic trading system and recorded complete decision trajectories, including portfolio states and executed trades.
We then apply \sys to perform profit--cost accounting and diagnostic analysis, allowing us to explain not only whether an agent is economically viable, but also why it succeeds or fails.

\subsection{Detailed Experimental Setup}
\label{app:exp-setup}

\stitle{Cost assumptions.}
Specifically, we account for four \textbf{cost components}:
(1) LLM-side: LLM token cost is computed from each model's usage and pricing from the official provider endpoints, and is adjusted by the task success rate (where \(\rho\) = 98\% as described in LLM provider metrics) to reflect the effective cost per successful completion.
(2) Trading-side: Commission fee is charged per execution, consistent with typical retail brokerage fees~\footnote{\url{https://www.interactivebrokers.com/en/pricing/commissions-stocks.php}}. Thus, we calculate the relevant cost based on that documentation.
% (3) Infra-side: Variable infrastructure cost is set to \$0.20 per run, with a fixed monthly data subscription fee (\eg \$100/month~\footnote{\url{https://www.alphavantage.co/premium/}}). 
\add{(3) Infra-side: Infrastructure overhead is estimated at \$0.20 per run, together with a fixed monthly data subscription fee (\eg \$100/month~\footnote{\url{https://www.alphavantage.co/premium/}}). Both are included in system-level deployment cost.}
(4) Stochastic: A stochastic cost term(\eg max. \$0.5 per run) is added.

\stitle{Profit attribution baselines.}
To complement the cost-side accounting, we construct two \textbf{profit-side baselines} for attribution analysis.
The first is a systematic exposure baseline, which uses the return of the S\&P 500 index over the same market window to capture broad market movement.
In implementation, the market benchmark is applied only to the portion of capital initially invested in risky assets, while any uninvested capital remains in cash~\cite{johnson2017performance,brinson1985measuring}. This preserves the agent's initial exposure decision and avoids implicitly assuming 100\% market exposure when part or all of the initial capital is held as cash. 

Consequently, the Market Effect may differ across backbone models if their initial invested fractions differ; a value of zero indicates no capital was initially exposed to risky assets.
The second is a static allocation baseline, which follows a buy-and-hold strategy by keeping the initial portfolio allocation unchanged in the trading period.

% These baselines allow us to distinguish profits to three parts as illustrated in section~\ref{sec:profit-attribution}.

% [Q3] Clarification of Market Effect Attribution
% We apologize that this implementation detail was not clearly stated in the paper. In our implementation, market exposure is applied only to the initially invested portion of the portfolio, while uninvested cash remains cash. This is because cash is itself part of the portfolio allocation. Applying the market benchmark to the full initial capital would implicitly assume 100% market exposure, even when the agent chooses to hold part or all of the capital in cash, and would therefore compare portfolios with different exposure bases [1,2].
% As a result, Market Effect varies across backbones because different models may produce different initial invested fractions under the same trading protocol. A value of 0.00 indicates that the agent held the initial capital in cash at the attribution starting point. Near-identical Market Effect or Asset Selection values can occur when different backbones produce the same or very similar initial invested allocations. We will revise Appendix B and Appendix D.1 to make this treatment explicit, align the formal definition with the implementation, and add a clearer table note for Table 1.

% [1] Johnson, D. (2017). Performance Attribution for Passive Strategies. Journal of Performance Measurement, 21(3).
% [2] Brinson, G. P., & Fachler, N. (1985). Measuring non-US. equity portfolio performance. The Journal of Portfolio Management, 11(3), 73-76.

\stitle{Trading setup}. The trading window is set to a two-month sideways market period from December 1, 2025 to January 30, 2026, starting with an initial cash of \$100{,}000 and trading a fixed universe of the most liquid 10 liquid U.S. equities. We accessed market data via licensed provider APIs/subscriptions and do not redistribute the raw data.
\add{Asset universe: NVDA, AAPL, MSFT, AMZN, GOOGL, GOOG, META, AVGO, TSLA, and WMT.
}

\subsection{Viability Across Capital Scale}
\label{app:rq2}

\begin{table*}[t]
\centering
\footnotesize
\resizebox{\linewidth}{!}{%
\begin{tabular}{@{ }l rrrr c rrrr@{ }}
\toprule
& \multicolumn{4}{c}{\textbf{DeepSeek-V3.2}} & & \multicolumn{4}{c}{\textbf{GPT-5.2}} \\ 
\cmidrule(lr){2-5} \cmidrule(lr){7-10}
\textbf{Metric} & \textbf{10k} & \textbf{50k} & \textbf{100k} & \textbf{500k}  & & \textbf{10k} & \textbf{50k} & \textbf{100k} & \textbf{500k}  \\ 
\midrule
\textbf{Gross Profit \textsuperscript{(a)}}      & -162.58 & -491.32 & -388.29 & -5,425.95  & & 303.15 & 1,278.89 & 505.76 & 2,022.49  \\ 
\quad Market Effect     & 16.38   & 829.56  & 1,351.45 & 7,715.41      & & 169.57 & 858.88  & 1,535.80 & 8,130.57 \\ 
\quad Asset Selection      & -114.91 & -598.14 & -1,500.44 & 2,686.92   & & 292.92 & 526.37  & -1,125.69 & -445.27 \\ 
\quad Timing Effect \textsuperscript{(1)}     & -64.05  & -722.73 & -239.30 & -15,828.28  & & -159.34 & -106.36 & 95.65 & -5,662.80  \\
\midrule
\textbf{Total Cost \textsuperscript{(b)}}        & 266.72  & 300.54  & 304.94  & 253.71      & & 228.13 & 237.66  & 258.79 & 242.46 \\ 
\quad Trading \textsuperscript{(2)}           & 32.00   & 53.00   & 63.00   & 19.18        & & 4.00   & 13.00   & 32.00 & 16.10  \\ 
\quad LLM \textsuperscript{(3)}          & 15.03   & 29.55   & 23.84   & 15.58    & & 4.52   & 7.37    & 9.03 & 6.72  \\ 
\quad Infra          & 208.20  & 208.20  & 208.20  & 208.20     & & 208.20 & 208.20  & 208.20 & 208.20  \\ 
\quad Stochastic \textsuperscript{(4)}            & 11.49   & 9.78    & 9.90    & 10.76    & & 11.41  & 9.09    & 9.56 & 11.45 \\ 
\midrule
\textbf{Net Profit \textsuperscript{(a-b)}}       & -429.30 & -791.85 & -693.23 & -5,679.66  & & 75.02  & 1,041.22 & 246.97 & 1,780.03  \\
\midrule
\textbf{Agentic Profit \textsuperscript{(1-2-3-4)}}      & -122.57 & -815.07 & -336.04 & -15,873.79  & & -179.27 & -135.82 & 45.06 & -5,697.07 \\ 
\bottomrule
\end{tabular}
}
\caption{Viability across Capital scale, ranging from 10k USD to 500k USD, comparing DeepSeek-V3.2 and GPT-5.2. All values are rounded to two decimal places from raw data.}
\label{tab:rq2-cap}
\end{table*}

\begin{table*}[t]
\centering
\footnotesize
\resizebox{0.65\linewidth}{!}{%
\begin{tabular}{l cc c cc}
\toprule
& \multicolumn{2}{c}{\textbf{DeepSeek-V3.2}} & & \multicolumn{2}{c}{\textbf{GPT-5.2}} \\ 
\cmidrule(lr){2-3} \cmidrule(lr){5-6}
\textbf{Decision Frequency} & \textbf{Hourly} & \textbf{Daily} & & \textbf{Hourly} & \textbf{Daily}  \\ 
\midrule
\textbf{Gross Profit \textsuperscript{(a)}}      & -906.86  & 197.23  & & -386.58 & -51.59  \\
\quad Market Effect     & 239.70   & 190.59  & & 241.95  & 216.59  \\ 
\quad Asset Selection     & -473.75  & 233.02  & & -169.13 & -73.62  \\ 
\quad Timing Effect \textsuperscript{(1)}     & -672.81  & -226.38 & & -459.40 & -194.56  \\
\midrule
\textbf{Total Cost \textsuperscript{(b)}}        & 315.38   & 167.38  & & 198.89  & 146.69  \\
\quad Trading \textsuperscript{(2)}           & 92.00    & 45.00  & & 19.00   & 32.00   \\
\quad LLM \textsuperscript{(3)}          & 63.41    & 13.17   & & 18.61   & 4.99    \\ 
\quad Infra          & 126.20   & 104.20  & & 126.20  & 104.20  \\
\quad Stochastic \textsuperscript{(4)}            & 33.78    & 5.01  &  & 35.08   & 5.50    \\
\midrule
\textbf{Net Profit \textsuperscript{(a-b)}}       & -1222.24 & 29.84  & & -585.46 & -198.28 \\
\midrule
\textbf{Agentic Profit \textsuperscript{(1-2-3-4)}}       & -861.99  & -289.56 & & -532.08 & -237.05  \\ 
\bottomrule
\end{tabular}
}
\caption{Viability across trading frequency.}
\label{tab:rq3-fre}
\end{table*}

Figure~\ref{fig:q2-capital} shows that increasing capital scale does not lead to a uniform improvement in economic viability.
Although larger capital may dilute fixed operating costs, the main variation in viability comes from profit-side components, especially asset selection and timing effects.

For GPT-5.2, system viability remains positive across all tested capital scales, although it does not increase monotonically with capital size.
Net profit rises from 75.02 at 10k to 1041.22 at 50k, decreases to 246.97 at 100k, and then increases again to 1780.03 at 500k.
However, agentic viability exhibits a different pattern.
It is negative at 10k and 50k, becomes slightly positive at 100k, and then drops sharply to -5697.07 at 500k.
Table~\ref{tab:rq2-cap} suggests that this deterioration at the largest tested scale is mainly driven by a strongly negative timing effect rather than by operating cost.
Although GPT-5.2 remains system viable at 500k, its dynamic intervention fails to generate positive active profit after accounting for decision-induced costs.

\begin{tcolorbox}[
title={Diagnosis sample: GPT-5.2 at 500k},
nobeforeafter,
breakable,
left=1mm,
right=1mm,
top=1mm,
bottom=1mm,
colback=white,
colframe=teal!50,
fontupper=\footnotesize,
halign=left
]
Top Recommended Actions:

Reduce execution latency and slippage (System/agent/tool architecture).  

   - Expected impact: rapidly recapture a large portion of the 3.98k opportunity cost and reduce timing erosion; could convert current fragile net profit into a robust positive.

   - Implementation idea: move to a trimmed inference pipeline (pruned prompt/context, async batching, prioritized decisions), colocate order gateway or use faster broker API paths, and implement pre‑validated order templates to eliminate per‑trade orchestration delays. Measure by reducing the average latency per trade below 5s and the slippage by half.  
\end{tcolorbox}

\vspace{9pt}
DeepSeek-V3.2 shows a consistently weaker pattern.
Both system and agentic viability remain negative across all capital scales.
The losses are moderate at 10k--100k but expand substantially at 500k, where system viability decreases to -5679.66, and agentic viability decreases to -15873.79.
Table~\ref{tab:rq2-cap} shows that this deterioration is not primarily caused by system cost, which remains small relative to profit variation.
Instead, the largest losses come from unfavorable profit components, especially the strongly negative timing effect at 500k.
These results suggest that capital scaling does more than dilute fixed costs.
For GPT-5.2, market exposure and asset selection remain positive in aggregate across scales, while timing varies non-monotonically and becomes strongly negative at 500k.
\vspace{9pt}

% The diagnosis, therefore, points to strategy-level failure rather than fee burden: DeepSeek-V3.2's market judgment and reallocation decisions introduce substantial capital losses as scale increases.

% \qi{todo: change the example}
% \begin{tcolorbox}[
% title={Diagnosis sample: DeepSeek-V3.2 at 1M},
% nobeforeafter,
% breakable,
% left=1mm,
% right=1mm,
% top=1mm,
% bottom=1mm,
% colback=white,
% colframe=teal!50,
% fontupper=\footnotesize,
% halign=left
% ]
% Primary failure mode is timing, including model forecasting and/or trading cadence, rather than fee burden. 
% Dynamic fees and token costs are immaterial relative to the timing loss: dynamic cost is \$67.52, with a cost-to-gross-profit ratio of approximately 1\%.
% The immediate focus should therefore be improving timing, model choice, trading cadence, and position-sizing rules; reducing LLM latency is secondary but may help recover execution-sensitive value.
% \end{tcolorbox}

\begin{tcolorbox}[
title={Diagnosis sample: DeepSeek-V3.2 at 500k},
nobeforeafter,
breakable,
left=1mm,
right=1mm,
top=1mm,
bottom=1mm,
colback=white,
colframe=teal!50,
fontupper=\footnotesize,
halign=left
]
  - Token intensity is large (239k input tokens/day) — this implies expensive and heavy LLM usage and long prompt/context windows, increasing compute latency and opportunity cost.
  
  - Distinction: selection skill is present (asset picks positive), but execution/system design (latency, LLM workflow) is producing missed fills and timing slippage. Market exposure (holding fast-rising assets) helped, but the system could not capture that upside because of when/how it traded.
\end{tcolorbox}
\newpage

% Overall, these results suggest that capital scaling mainly acts as an amplifier of backbone-specific trading behavior.
% For GPT-5.2, larger capital can amplify profitable timing decisions and offset losses that would have occurred under the initial buy-and-hold allocation.
% For DeepSeek-V3.2, larger capital instead magnifies strategy-level losses caused by poor asset selection and ineffective timing.
% Therefore, larger capital can dilute fixed costs, but cost dilution alone is insufficient to ensure economic viability; the effect of scaling depends on whether the agentic strategy can generate positive active profit after costs.

For DeepSeek-V3.2, the large loss at 500k is dominated by negative timing despite positive asset selection.
Thus, scaling can substantially change decision-attributed outcomes, and cost dilution alone is insufficient to ensure economic viability.

\subsection{Viability Across Trading Frequency}
\label{app:rq3}

Figure~\ref{fig:q3-frequency} shows that higher decision frequency does not improve viability. 
Compared with daily trading, hourly trading reduces gross profit by 1104.09 for DeepSeek-V3.2 and 334.99 for GPT-5.2, while the additional total cost accounts for only 11.8\% and 13.5\% of the corresponding net-profit deterioration.
This indicates that the main failure mode is not fee burden, but degradation in trading decisions.

The attribution results, shown in Table~\ref{tab:rq3-fre}, reveal that the degradation is concentrated in the active components.

For both backbones, hourly trading yields more negative timing effects than daily trading, suggesting that more frequent decision points add timing errors rather than stable short-term opportunities.

For DeepSeek-V3.2, the failure is more severe: hourly trading worsens both asset selection and timing effects, as frequent reallocation affects not only when the agent trades but also what it holds.
Beyond the aggregate profit--cost accounting, we apply \sys to explain why higher decision frequency fails to improve viability.
The diagnosis shows that the hourly setting does not merely increase operating cost, but also changes the quality of the active decisions.
Thus, the hourly setting exposes an important issue: when the predictive signal is weak, increasing decision frequency amplifies noise, turnover, and mistimed reallocations instead of improving economic viability.

\begin{tcolorbox}[
title={Diagnosis sample: DeepSeek-V3.2 under hourly trading},
nobeforeafter,
breakable,
left=1mm,
right=1mm,
top=1mm,
bottom=1mm,
colback=white,
colframe=teal!50,
fontupper=\footnotesize,
halign=left
]
Trading cadence is likely too aggressive relative to predictive signal quality: 190 trades/month across multiple assets combined with negative timing indicates frequency is not justified by signal edge.
Portfolio construction also contributed: initial holdings and sizing choices (see initial holdings versus final positions) show nontrivial exposures that did not capture market moves.
\end{tcolorbox}

\subsection{Viability Across System Architecture}
\label{app:rq4}

\stitle{Chain-of-Thought baseline.}
The CoT baseline uses the same market data and portfolio context as the agentic system, but removes tool interaction and multi-step external execution.
For each trading date, the system first retrieves the current portfolio, daily open prices, previous open and close prices, and recent news for the watchlist.

It then constructs a single prompt that asks the LLM to analyze all tickers, provide a short portfolio-level synthesis, and output a final trading decision in a structured JSON format.
Only the final JSON block is used for trade execution, while the preceding step-by-step analysis is logged as the model's reasoning trace.

Table~\ref{tab:rq4-system} compares economic viability across different system architectures under the same initial capital of \$100{,}000.
The results show that the economic value of an architecture does not depend simply on whether it is cheaper or more complex, but on whether additional reasoning and coordination can be converted into better asset selection and timing decisions.

While CoT has the lowest total cost for both backbones, it fails to achieve the best economic outcome, proving that low operating costs alone do not guarantee viability. Weak reasoning structures still result in poor asset selection and ineffective reallocation despite lower expenses. This flaw is especially clear in DeepSeek-V3.2, where CoT caused a strongly negative timing effect and the worst overall agentic profit.

In contrast, DeepFund yields more stable gains across both backbones. It is the only architecture to achieve both positive net and agentic profit for DeepSeek-V3.2, while also capturing the highest net and agentic profit for GPT-5.2. Crucially, DeepFund does not have the lowest total cost. This proves its competitive advantage stems entirely from superior profit generation rather than cost reduction.

The attribution results further suggest that DeepFund's advantage is mainly associated with timing.
It produces positive timing effects for both backbones, indicating that the architecture does not merely benefit from market exposure, but it actually also creates active value through dynamic reallocation.
For retail investors interested in developing agentic trading systems, this suggests that the key challenge is not only to reduce inference cost or add more reasoning steps, but also to monitor complete trading trajectories and to verify whether reasoning cost is actually converted into better investment outcomes.\\

\begin{tcolorbox}[
title={Diagnosis sample: DeepSeek-V3.2 under CoT},
nobeforeafter,
breakable,
left=1mm,
right=1mm,
top=1mm,
bottom=1mm,
colback=white,
colframe=teal!50,
fontupper=\footnotesize,
halign=left
]
  Dynamic cost accumulation is tiny relative to portfolio swings, so cumulative cost lines are almost flat while portfolio value moves dominate.
  The run exhibits path dependency driven by timing decisions: the timing effect (-3,214.75) shows losses are concentrated in when trades were executed, not in cost build-up.
  
- Fragility of profitability:

  - Given small cost base and positive selection when capital is invested, profitability is fragile but fixable — i.e., the strategy can produce positive selection returns if timing is corrected or exposure windows are altered.
  
  - However, until timing decisions improve, increasing capital or leaving the static subscription in place risks repeated net losses.
\end{tcolorbox}

\subsection{Sensitivity in Market Regime}
\label{app:regime}

% \begin{table}[htbp]
% \centering
% \footnotesize
% % \small
% \resizebox{\linewidth}{!}{%
% \begin{tabular}{l ccc c ccc}
% \toprule
% & \multicolumn{3}{c}{\textbf{DeepSeek-V3.2}} & & \multicolumn{3}{c}{\textbf{GPT-5.2}} \\ 
% \cmidrule(lr){2-4} \cmidrule(lr){6-8}
% \textbf{Regime} & Bearish & Bullish & Sideway & & Bearish & Bullish & Sideway \\ 
% \midrule
% \textbf{Profit} & -21770.56 & 11249.93 & -388.29 & & -27966.93 & 14887.99 & 505.76 \\ 
% \midrule
% \textbf{Cost} & 297.25 & 285.26 & 299.93 & & 237.73 & 289.10 & 258.26 \\ 
% \quad Trading & 239.00 & 229.00 & 263.00 & & 212.00 & 258.00 & 232.00 \\ 
% \quad LLM & 39.21 & 35.64 & 18.26 & & 7.53 & 9.74 & 8.85 \\ 
% \quad Infra & 7.80 & 9.40 & 8.20 & & 7.80 & 9.40 & 8.20 \\ 
% \quad Stochastic & 11.24 & 11.91 & 10.47 & & 10.40 & 11.96 & 9.21 \\ 
% \midrule
% \textbf{Net Profit} & \textbf{-22067.81} & \textbf{10963.98} & \textbf{-688.23} & & \textbf{-28204.66} & \textbf{14598.89} & \textbf{247.5} \\ 
% \bottomrule
% \end{tabular}
% }
% \caption{Viability across market regime: profit and cost breakdown with an initial cash of \$100{,}000.}
% \label{tab:market-regime}
% \end{table}

\begin{table*}[t] 
\centering
\small
\resizebox{0.9\linewidth}{!}{%
\begin{tabular}{l ccc c ccc}
\toprule
& \multicolumn{3}{c}{\textbf{DeepSeek-V3.2}} & & \multicolumn{3}{c}{\textbf{GPT-5.2}} \\ 
\cmidrule(lr){2-4} \cmidrule(lr){6-8}
\textbf{Metric} & \textbf{Bearish} & \textbf{Sideways} & \textbf{Bullish} & & \textbf{Bearish} & \textbf{Sideway} & \textbf{Bullish} \\ 
\midrule
\textbf{Gross Profit \textsuperscript{(a)}} & -21770.56 & -388.30 & 11249.93 & & -27966.93 & 505.76 & 14887.99 \\
\quad Market Effect & -2488.54 & 1351.45 & 755.35 & & -16570.14 & 1535.80 & 4294.88 \\
\quad Asset Selection & -2328.65 & -1500.44 & 272.15 & & -10920.83 & -1125.69 & 6735.38 \\
\quad Timing Effect \textsuperscript{(1)} & -16953.37 & -239.30 & 10222.43 & & -475.96 & 95.65 & 3857.73 \\
\midrule
\textbf{Total Cost \textsuperscript{(b)}} & 281.89 & 304.94 & 274.08 & & 236.11 & 258.79 & 286.83 \\
\quad Trading \textsuperscript{(2)} & 39.00 & 63.00 & 29.00 & & 12.00 & 32.00 & 58.00 \\
\quad LLM \textsuperscript{(3)} & 26.24 & 23.84 & 23.85 & & 7.53 & 9.03 & 9.74 \\
\quad Infra & 207.80 & 208.20 & 209.40 & & 207.80 & 208.20 & 209.40 \\
\quad Stochastic \textsuperscript{(4)} & 8.84 & 9.90 & 11.83 & & 8.77 & 9.56 & 9.69 \\
\midrule
\textbf{Net Profit \textsuperscript{(a-b)}} & -22052.45 & -693.23 & 10975.85 & & -28203.04 & 246.97 & 14601.16 \\
\midrule
\textbf{Agentic Profit \textsuperscript{(1-2-3-4)}} & -17027.46 & -336.04 & 10157.75 & & -504.27 & 45.06 & 3780.30 \\
\bottomrule
\end{tabular}
}
\caption{Viability across market conditions. Cumulative profit and cost breakdown with an initial cash of \$100{,}000.}
\label{tab:market-regime}
\end{table*}

Market regime is typically discussed for its impact on returns~\cite{ang2012regime,dong2025large}. 
For agentic trading systems, regime shifts may also affect viability by changing both profit opportunities and deployment frictions. 
For example, liquidity and volatility can alter execution costs such as spread and slippage, while different market conditions may induce different trading intensity and runtime workload. 
Thus, before attributing net-profit changes to model or system design alone, we test whether profit--cost viability is sensitive to market regimes.

In order to isolate temporal and regime effects, we conduct several additional experiments under bearish (2025-02-10 to 2025-04-07), sideways (2025-12-01 to 2026-01-30), and bullish (2025-06-02 to 2025-08-08) conditions. 
The results are shown in Table~\ref{tab:market-regime} , which shows that profit varies much more sharply than cost across regimes. 
For both DeepSeek-V3.2 and GPT-5.2, total costs remain within a relatively narrow range, whereas gross profit, net profit, and agentic profit change substantially with market conditions.

In the bearish regime, both models are system-non-viable, but the failure mechanisms differ. 
DeepSeek-V3.2 suffers from a large negative timing effect, indicating that dynamic reallocation amplifies downside exposure. 
GPT-5.2, by contrast, is mainly hurt by negative market exposure and asset selection, while its timing loss is relatively small. 
In the sideways regime, GPT-5.2 achieves modest positive net profit and agentic profit, whereas DeepSeek-V3.2 remains negative, suggesting that the same market condition can still expose backbone-specific differences in timing and execution quality. 
In the bullish regime, both models become strongly profitable, but through different profit sources: DeepSeek-V3.2 is dominated by timing gains, while GPT-5.2 benefits from both asset selection and timing.
Overall, the market regime mainly determines whether LLM-mediated decisions can yield profit.
This result reinforces the need for profit--cost diagnosis: similar cost levels can correspond to very different viability outcomes because market exposure, asset selection, and timing respond differently across regimes.

\subsection{Complementary Financial Performance Metrics}
\label{app:financial_metrics}

We additionally report conventional financial metrics as complementary diagnostics for the evaluated trading configurations. Agentic viability is not intended to replace these standard measures: conventional metrics characterize the realized performance and risk profile of the overall portfolio, whereas our analysis focuses on whether the dynamic LLM-mediated layer generates sufficient incremental value to justify its costs.

\paragraph{Total Return.}
Total return measures the cumulative change in portfolio value over the evaluation period:

$$
R =
\frac{V_T - V_0}{V_0},
$$

\noindent where \(V_0\) and \(V_T\) denote the initial and final portfolio values, respectively.

\paragraph{Sharpe Ratio.}
The annualized Sharpe ratio is computed from daily portfolio returns:

$$
S =
\frac{\bar{r} - r_f}{\sigma_r}
\sqrt{252},
$$

where \(\bar{r}\) is the average daily return, \(r_f\) is the daily risk-free rate, and \(\sigma_r\) is the standard deviation of daily returns.

\paragraph{Maximum Drawdown.}
Maximum drawdown measures the largest decline from a historical portfolio-value peak:

$$
\mathrm{MDD}
=
\max_t
\left(
\frac{V_t^{\max} - V_t}{V_t^{\max}}
\right),
$$

\noindent where

$$
V_t^{\max} = \max_{\tau \leq t} V_\tau .
$$

\paragraph{Turnover.}
Turnover measures the amount of trading relative to portfolio value:

$$
\mathrm{TO}
=
\frac{1}{T}
\sum_{t=1}^{T}
\frac{A_t}{V_t},
$$

\noindent where \(A_t\) denotes the total traded amount at trading step \(t\), \(V_t\) is the portfolio value, and \(T\) is the number of trading steps.

\begin{table}[t] 
\centering
\small
\resizebox{\linewidth}{!}{
\begin{tabular}{l cccc}
\toprule
\bf Model             & \bf Total & \bf Sharpe & \bf Max & \bf Turn- \\
           &  \bf Return(\%) & \bf S& DD(\%) & \bf over \\
\midrule
Qwen3-Max         & -0.05      & -5.81  & 0.22 & 0.08     \\
Mistral-Large-3   & 0.78       & 0.13   & 4.71 & 5.06     \\
MiniMax-M2.1      & -1.47      & -0.70  & 6.73 & 2.21     \\
Llama-4-scout     & -1.02     & -0.71  & 4.9 & 0.92     \\
Kimi-K2           & -1.98      & -1.06  & 5.84 & 2.13     \\
GPT-5.2           & 0.51       & 0.02   & 4.79 & 1.16     \\
GLM-4.7           & -2.55     & -1.30  & 6.44 & 5.92     \\
Gemini 3 Flash    & -0.98     & -0.55  & 7.06 & 5.67     \\
DeepSeek-V3.2     & -0.39     & -0.42  & 4.96 & 2.37     \\
Claude Sonnet 4.5 & 0.52       & 0.03   & 5.69 & 1.33    \\
\bottomrule
\end{tabular}
}
\caption{Standard financial metrics for the model comparison experiment with \$100K initial capital and daily trading frequency (time window: Dec. 2025–Jan. 2026).}
\label{tab:app-1}
\end{table}

% \begin{table}[t] 
% \centering
% \small
% \resizebox{\linewidth}{!}{%
% \begin{tabular}{llcccc}
% \toprule
% Model         & Capital & Total Return(\%) & Sharpe & Max DD(\%) & Turnover \\
% \midrule
% DeepSeek-V3.2 & \$10k   & -1.63      & -0.98  & 5.82 & 2.39     \\
% DeepSeek-V3.2 & \$50k   & -0.98      & -0.67  & 4.89 & 1.97     \\
% DeepSeek-V3.2 & \$100k  & -0.39     & -0.42  & 4.9 & 2.37     \\
% DeepSeek-V3.2 & \$500k  & -1.09      & -0.97  & 5.49 & 1.38     \\
% GPT-5.2       & \$10k   & 3.03       & 1.04   & 5.41 & 0.98     \\
% GPT-5.2       & \$50k   & 2.56       & 0.82   & 5.06 & 1.39     \\
% GPT-5.2       & \$100k  & 0.51       & 0.02   & 4.79 & 1.16     \\
% GPT-5.2       & \$500k  & 0.4        & -0.02  & 5.39 & 1.13    \\
% \bottomrule
% \end{tabular}
% }
% \caption{Standard financial metrics for the capital scale experiment with daily trading frequency (time window: Dec 25 - Jan 26).}
% \label{tab:app-2}
% \end{table}

\begin{table}[t] 
\centering
\small
\resizebox{\linewidth}{!}{%
\begin{tabular}{@{ }llcccc@{ }}
\toprule
\bf Model         & \bf Capital & \bf Total & \bf Sharpe & \bf Max & \bf Turn- \\
 &  & \bf Return(\%) & & \bf DD(\%) & \bf over \\
\midrule
\multirow{4}{*}{DeepSeek-V3.2}
 & \$10k   & -1.63      & -0.98  & 5.82 & 2.39     \\
 & \$50k   & -0.98      & -0.67  & 4.89 & 1.97     \\
 & \$100k  & -0.39     & -0.42  & 4.9 & 2.37     \\
 & \$500k  & -1.09      & -0.97  & 5.49 & 1.38     \\
 \midrule
\multirow{4}{*}{GPT-5.2}
 & \$10k   & 3.03       & 1.04   & 5.41 & 0.98     \\
 & \$50k   & 2.56       & 0.82   & 5.06 & 1.39     \\
 & \$100k  & 0.51       & 0.02   & 4.79 & 1.16     \\
 & \$500k  & 0.4        & -0.02  & 5.39 & 1.13    \\
\bottomrule
\end{tabular}
}
\caption{Standard financial metrics for the capital scale experiment with daily trading frequency (time window: Dec. 2025–Jan. 2026).}
\label{tab:app-2}
\end{table}

% \begin{table}[t] 
% \centering
% \small
% \resizebox{\linewidth}{!}{%
% \begin{tabular}{l ccccc}
% \toprule
% Model         & Freq   & Total Return(\%) & Sharpe & Max DD(\%) & Turnover \\
% \midrule
% DeepSeek-V3.2 & daily  & 0.2        & -0.06  & 3.77 & 1.93     \\
% DeepSeek-V3.2 & hourly & -1.55      & -1.56  & 3.51 & 3.35     \\
% GPT-5.2       & daily  & -0.05      & -0.28  & 3.68 & 1.16     \\
% GPT-5.2       & hourly & -1.66      & -1.32  & 3.58 & 1.35    \\
% \bottomrule
% \end{tabular}
% }
% \caption{Standard financial metrics for the trading frequency experiment (time window: Dec 25 - Jan 26).}
% \label{tab:app-3}
% \end{table}

\begin{table}[t] 
\centering
\small
\resizebox{\linewidth}{!}{%
\begin{tabular}{@{ }l ccccc@{ }}
\toprule
\bf Model         & \bf Freq   & \bf Total & \bf Sharpe & \bf Max & \bf Turn- \\
         &   &Return(\%) &  & \bf DD(\%) & \bf over \\
\midrule
\multirow{2}{*}{DeepSeek-V3.2}
 & daily  & 0.2        & -0.06  & 3.77 & 1.93     \\
 & hourly & -0.91      & -0.62 & 4.12 & 3.36     \\
 % & hourly & -1.55      & -1.56  & 3.51 & 3.35     \\
\midrule
\multirow{2}{*}{GPT-5.2}
 & daily  & -0.05      & -0.28  & 3.68 & 1.16     \\
 % & hourly & -1.66      & -1.32  & 3.58 & 1.35    \\
  & hourly & -0.39      & -0.05  & 5.86 & 1.34    \\
\bottomrule
\end{tabular}
}
\caption{Standard financial metrics for the trading frequency experiment (time window: Dec. 2025).}
\label{tab:app-3}
\end{table}

% \begin{table}[t] 
% \centering
% \small
% \resizebox{\linewidth}{!}{%
% \begin{tabular}{l ccccc}
% \toprule
% Model         & Variant  & Total Return(\%) & Sharpe & Max DD(\%) & Turnover \\
% \midrule
% DeepSeek-V3.2 & CoT      & -1.35      & -0.75  & 5.18 & 2.83     \\
% DeepSeek-V3.2 & DeepFund & 0.63       & 0.01   & 2.67 & 6.87     \\
% GPT-5.2       & CoT      & 2.75       & 0.65   & 7.33 & 2.43     \\
% GPT-5.2       & DeepFund & 1.07       & 0.27   & 3.44 & 5.58     \\
% \bottomrule
% \end{tabular}
% }
% \caption{Standard financial metrics for the system complexity experiment (time window: Dec 25 - Jan 26).}
% \label{tab:app-4}
% \end{table}

\begin{table}[t] 
\centering
\small
\resizebox{\linewidth}{!}{%
\begin{tabular}{@{ }l ccccc@{ }}
\toprule
\bf Model         & \bf Variant  & \bf Total & \bf Sharpe & \bf Max & \bf Turn- \\
     &   & \bf Return(\%) &  & \bf DD(\%) & \bf over \\
\midrule
\multirow{2}{*}{DeepSeek-V3.2}
 & CoT      & -1.24      & -0.70  & 5.18 & 2.83     \\
 & DeepFund & 0.61       & -0.02   & 2.65 & 6.87     \\
\midrule
\multirow{2}{*}{GPT-5.2}
 & CoT      & 0.34       & -0.04   & 4.55 & 1.82     \\
 & DeepFund & 0.88       & 0.06   & 3.46 & 5.58     \\
\bottomrule
\end{tabular}
}
\caption{Standard financial metrics for the system complexity experiment (time window: Dec. 2025–Jan. 2026).}
\label{tab:app-4}
\end{table}

% \begin{table}[t] 
% \centering
% \small
% \resizebox{\linewidth}{!}{%
% \begin{tabular}{ll ccccc}
% \toprule
% Model         & Window                & Condition & Total Return(\%) & Sharpe & Max DD(\%)  & Turnover \\
% DeepSeek-V3.2 & 2025-02 $\sim$2025-04 & bearish   & -21.41     & -5.74  & 21.90 & 1.49     \\
% DeepSeek-V3.2 & 2025-06 $\sim$2025-08 & bullish   & 10.89       & 3.68   & 3.49  & 1.4      \\
% GPT-5.2       & 2025-02 $\sim$2025-04 & bearish   & -24.77     & -5.55  & 25.46 & 1.18     \\
% GPT-5.2       & 2025-06 $\sim$2025-08 & bullish   & 14.89       & 4.56   & 3.67  & 1.98  \\
% \bottomrule
% \end{tabular}
% }
% \caption{Standard financial metrics for the market regime experiment.}
% \label{tab:app-5}
% \end{table}

\begin{table*}[t] 
\centering
\small
\resizebox{0.9\linewidth}{!}{%
\begin{tabular}{ll ccccc}
\toprule
\bf Model         & \bf Window                & \bf Condition & \bf Total Return (\%) & \bf Sharpe & \bf Max DD(\%)  & \bf Turnover \\
\midrule
\multirow{2}{*}{DeepSeek-V3.2}
 & 2025-02 $\sim$2025-04 & bearish   & -21.77     & -5.72  & 22.21 & 1.49     \\
 & 2025-06 $\sim$2025-08 & bullish   & 11.25       & 3.76   & 3.51  & 1.4      \\
\midrule
\multirow{2}{*}{GPT-5.2}
 & 2025-02 $\sim$2025-04 & bearish   & -27.97     & -5.96  & 28.7 & 1.20     \\
 & 2025-06 $\sim$2025-08 & bullish   & 14.89       & 4.56   & 3.67  & 1.98  \\
\bottomrule
\end{tabular}
}
\caption{Standard financial metrics for the market regime experiment.}
\label{tab:app-5}
\end{table*}

Tables~\ref{tab:app-1}, \ref{tab:app-2}, \ref{tab:app-3}, \ref{tab:app-4}, and~\ref{tab:app-5} report total return, Sharpe ratio, maximum drawdown, and turnover for the experimental settings in Section~\ref{sec:exp-o1-o4} and Appendix~\ref{app:regime}.

\subsection{\add{Extended Evaluation}}
\label{app:extended-exp}

\begin{table*}[t!]
\centering
\small
\resizebox{\textwidth}{!}{%
\begin{tabular}{lrrrrrrrrrrr}
\toprule
\bf Model 
& \bf Total
& \bf Sharpe 
& \bf Max DD
& \bf Turnover 
& \bf Gross
& \bf Market
& \bf Selection
& \bf Timing
& \bf Total 
& \bf Net
& \bf Agentic\\
\bf 
& \bf Return (\%) 
& \bf 
& \bf (\%) 
& \bf 
& \bf (\$) 
& \bf 
& \bf 
& \bf 
& \bf Cost (\$) 
& \bf Profit (\$) 
& \bf Profit (\$) \\
\midrule
GPT-5.4
& 5.08
& 0.37
& 14.75
& 18.64
& 5,080
& 8,754
& $-9,328$
& 5,655
& 873
& 4,207
& 5,407 \\

DeepSeek-V4-Flash
& $-6.23$
& $-0.61$
& 18.28
& 13.77
& $-6,228$
& 9,192
& $-9,558$
& $-5,862$
& 1,068
& $-7,296$
& $-6,305$ \\

Gemini 3.5 Flash
& $-11.31$
& $-0.62$
& 28.35
& 163.93
& $-11,310$
& 9,290
& $-18,551$
& $-2,048$
& 1,040
& $-12,350$
& $-2,464$ \\
\bottomrule
\end{tabular}%
}
\caption{Extended evaluation of three representative frontier backbone models. All agents start with \$100K initial capital and make daily trading decisions from January to July 2026. SPY returns 9.36\% over the same period.}
\label{tab:six-month}
\end{table*}

To further examine whether the findings are specific to the original two-month evaluation window, we conduct an extended six-month experiment using three representative frontier backbone models: GPT-5.4\cite{openai2026gpt54}, DeepSeek-V4-Flash\cite{deepseek2026v4}, and Gemini 3.5 Flash\cite{kavukcuoglu2026gemini35}. Each agent starts with \$100K initial capital and makes daily trading decisions from January 1st to July 1st 2026(SPY returns 9.36\%, 124 trading days).

Table~\ref{tab:six-month} reports both conventional financial metrics and TradeLens attribution results. GPT-5.4 is the only model that achieves a positive net return, while all agents underperform the SPY benchmark. Across models, the positive market component is offset by negative selection value, while timing differs substantially across agents: GPT-5.4 partially recovers selection losses through positive timing, whereas DeepSeek-V4-Flash and Gemini 3.5 Flash incur additional losses from dynamic trading. 

These results suggest that extending the evaluation horizon does not eliminate the central viability issue: realized portfolio performance depends not only on market exposure, but also on whether dynamic agentic decisions generate sufficient incremental value relative to their associated costs.

\section{Deployment User Study}

\label{deployment}

This part jointly evaluates deployability from both technical and user-facing perspectives: whether \sys can provide forward-looking deployment diagnostics, and whether practitioners find such diagnostics useful for analyzing and revising agentic trading systems.

\subsection{Forward Diagnostic Analysis}
\label{app:forward}

\begin{figure}
    \centering
    \includegraphics[width=\linewidth]{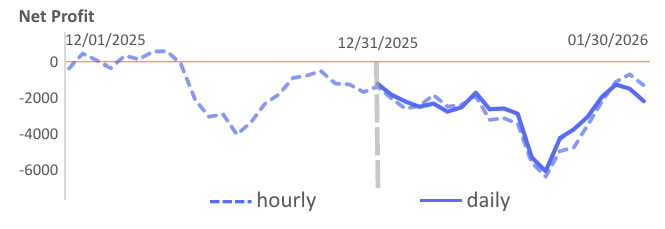}
    \caption{Forward diagnostic analysis of trading frequency. Cumulative net profit of DeepSeek-V3.2 under hourly (dashed line) and daily (solid line) trading frequencies. The vertical gray line marks the transition date.}
    \label{fig:rq5-forward}
\end{figure}

\stitle{Diagnostic finding in December 2025.}
Under the hourly configuration, the testing system starts with \$100,000 and ends in December with a gross loss of approximately \$900, while incurring about \$300 in total system cost. 

Fixed data subscription fees account for roughly \$100, and the main dynamic costs come from trading-side and LLM-side usage. The diagnostic report further identifies excessive token usage per decision and concentrated capital utilization. Based on these accounting signals, \sys flags trading frequency as a major cost driver and recommends reducing decision frequency as a high-priority adjustment.

\stitle{Forward evaluation in January 2026.}
We evaluate the frequency-reduction suggestion in a later market window by comparing the daily configuration with hourly baseline. In January, the daily configuration reduces total system cost from approximately \$610 under the hourly baseline to about \$124, and achieves higher net profit in early January~(Figure~\ref{fig:rq5-forward}). However, net profit remains negative under both configurations. A reversal occurs on January 28, when the daily configuration abstains from trading under unfavorable market conditions, whereas the hourly configuration continues active trading and partially mitigates losses. By the end of January, the hourly baseline records a smaller net loss than the reduced-frequency configuration.

These results show that the December diagnosis correctly identifies trading frequency as an important cost driver, but cost reduction alone does not guarantee later-window viability. When gross returns are constrained by market conditions and strategy-level edge, lowering decision frequency can reduce cost while still failing to improve final net profit. Thus, the forward analysis supports the role of \sys as a diagnostic tool: it explains why a system fails to become viable, rather than assuming that a cost-saving suggestion will always improve performance.

\subsection{Practitioner Feedback}
\label{app:deploy}

\begin{table}[t]
\centering
\footnotesize
\resizebox{\linewidth}{!}{%
\begin{tabular}{@{ }lcccc@{ }}
\toprule
\textbf{ID}  & \textbf{Background} & \textbf{Trading Exp.} & \textbf{AI Exp.} \\
\midrule
P1   & Undergraduate Student      & 1--3 yrs  & Low \\
P2   & Graduate Student           & 1--3 yrs  & Medium \\
P3   & Individual Investor        & 3--5 yrs  & Medium \\
P4   & Individual Investor        & 3--5 yrs  & Low \\
P5   & Finance Practitioner       & 5+ yrs    & Medium \\
P6   & Graduate Student           & 1--3 yrs  & Low \\
P7   & Data Analyst               & 3--5 yrs  & Medium \\
P8   & AI System Developer        & 5+ yrs    & High \\
P9   & Individual Investor        & 1--3 yrs  & Medium \\
P10  & Software Engineer          & 3--5 yrs  & Low \\
P11  & Finance Practitioner       & 5+ yrs    & Medium \\
P12  & Undergraduate Student      & 1--3 yrs  & Low \\
P13  & Retail Trading Enthusiast  & 3--5 yrs  & Medium \\
\bottomrule
\end{tabular}%
}
\caption{Participant demographics for the deployment-oriented usability assessment.}
\label{tab:demographics}
\end{table}

% \begin{figure}
%     \centering
%     \includegraphics[width=0.9\linewidth]{figures/appendix/appendix-deployment-example.pdf}
%     \caption{Example deployment cases and feedback from human experts for different retail trading setups.}
%     \label{fig:feedback}
% \end{figure}

% To complement the forward diagnostic analysis, we collected practitioner feedback on whether \sys helps users interpret the deployability of private agentic trading systems. We recruited 13 retail traders who integrated the toolkit into their own agentic trading workflows while keeping their strategies, assets, and portfolio details private. After a brief usage guide, participants used \sys to audit profit sources, cost drivers, net profit, and viability status, and then provided qualitative feedback through a 5-point Likert questionnaire.
% %
% We also invited five external reviewers with financial or AI engineering backgrounds to assess the generated reports. The financial reviewers focused on whether the reports provided feasible and actionable interpretations, while the AI engineering reviewers assessed whether the accounting results and diagnostic explanations were clear and traceable. 
% %
% Participant demographics are summarized in Table~\ref{tab:demographics}.

\stitle{Participants.}
In order to complement the forward diagnostic analysis, we further collected practitioner's feedback on whether \sys helps users interpret the deployability of private agentic trading systems.
We recruited a total of thirteen retail traders who used \sys in their own agentic trading workflows while keeping their strategies, asset universes, trading logs, and portfolio details private.
The participants were recruited through social networks, and a startup company specializing in financial AI, who agreed to support this effort. Several employees of the company voluntarily participated in the experiment and were allowed to withdraw at any time. To preserve their anonymity, we do not disclose the company name. In this collaboration, we collected only participants' relevant professional experience and feedback, and did not collect any personally identifiable information.
% We also invited five external reviewers with financial or AI engineering backgrounds to assess representative diagnostic reports. Financial reviewers focused on whether the reports provided feasible and actionable trading interpretations, while AI engineering reviewers assessed whether the accounting results and diagnostic explanations were clear and traceable.
Participant demographics are summarized in Table~\ref{tab:demographics}.

\stitle{Procedure.}
In order to ensure valid participation, we carefully explained the purpose of the data collection and how the collected data would be used afterwards, and we obtained informed consent from all participants before any data collection has started.
The participants then received a brief usage guide explaining the required inputs, profit--cost metrics, and report structure.
They then used \sys in order to audit profit sources, cost drivers, net profit, system viability, and agentic viability for their own runs.
After using the toolkit, they completed a 5-point Likert questionnaire and provided open-ended comments.
The questionnaire covered the following three dimensions:
(i) \emph{profit--cost diagnosis}, including whether the report helped users understand viability, distinguish market exposure, asset selection, and timing, identify dominant cost drivers, and separate weak decision value from high cost,
(ii) \emph{diagnostic actionability}, including whether the report supported revision decisions such as model choice, trading frequency, architecture, or execution adjustment, and
(iii) \emph{usability and integration}, including whether the outputs were understandable and whether the integration effort was acceptable.
%

% add results

\stitle{Main feedback.}
Overall, the participants generally found \sys to be quite useful for diagnosing why a private trading agent was or was not deployable, instead of merely reporting its final profit.
They emphasized that profit and cost should be interpreted jointly, since poor net performance may result from different underscoring mechanisms, including weak asset selection, mistimed reallocation, excessive trading friction, high LLM usage, or fixed data and infrastructure fees.
The participants further found the distinction between system viability and agentic viability helpful, because it separates systems that merely benefit from market exposure from those whose dynamic decisions add incremental value.

From a developer's perspective, participants highlighted three practical benefits.
First, the reports made debugging more traceable by linking failure modes to trading records, runtime traces, and cost components.
Second, the configuration-level recommendations were actionable, such as reducing trading frequency, limiting redundant model calls, or improving execution design.
Third, the reports helped prioritize whether a failure should be addressed through operational tuning or deeper architectural revision.

Participants also noted several limitations.
They requested more concrete strategy-level guidance after failure modes are identified, especially for asset selection, timing adjustment, and position control.
Some participants suggested making benchmark choice, execution-cost assumptions, and fixed infrastructure fees more explicit, so that the diagnosis can be better adapted to different deployment settings.
These results support \sys as a practical tool for assessing whether LLM-mediated decisions convert induced costs into deployable trading value under realistic private deployment constraints.

\stitle{Ethics statement.}
This feedback study was reviewed and approved by the institutional ethics committee of our university.
All participants were informed of the purpose of the study before providing feedback.
Participation was voluntary, and participants could withdraw at any time.
The collected feedback was used only for research purposes, and all demographic information was anonymized and reported in aggregate form.

\end{document}